\documentclass[conference]{IEEEtran}
\usepackage{times}

\usepackage[numbers]{natbib}
\usepackage{multicol}
\usepackage[bookmarks=true]{hyperref}
\usepackage{algorithm}
\usepackage{algpseudocode}
\usepackage{graphicx}
\usepackage{amsmath}
\usepackage{xspace}
\usepackage{subcaption}
\usepackage{multirow}
\usepackage{booktabs}
\usepackage{tabularx}

\renewcommand{\eqref}[1]{(\ref{#1})}
\newcommand{\tabref}[1]{Table~\ref{#1}}
\newcommand{\subfig}[1]{\textit{#1}}
\newcommand{\figref}[1]{Fig.~\ref{#1}}

\newcommand{\ie}{\textrm{i.e.}}
\newcommand{\eg}{\textrm{e.g.}}
\newcommand{\etc}{\textrm{etc.}}

\DeclareMathOperator*{\argmax}{arg\,max}

\newcommand{\NRP}{NRP\xspace}
\newcommand{\NRPRRT}{NRP\xspace}
\newcommand{\NRPIRRT}{NRP*\xspace}
\newcommand{\FIRE}{FIRE\xspace}
\newcommand{\FIREStar}{FIRE*\xspace}

\newcommand{\Wglobal}{\ensuremath{W^{global}}\xspace}
\newcommand{\Wlocal}{\ensuremath{W^{local}}\xspace}
\newcommand{\qLocalStart}{\ensuremath{q_{s}^{local}}\xspace}
\newcommand{\qLocalGoal}{\ensuremath{q_{g}^{local}}\xspace}
\newcommand{\qstart}{\ensuremath{q_{s}}\xspace}
\newcommand{\qgoal}{\ensuremath{q_{g}}\xspace}
\newcommand{\qcur}{\ensuremath{q_{cur}}\xspace}
\newcommand{\qtarget}{\ensuremath{q_{target}}\xspace}

\newcommand{\solPathOptimal}{\ensuremath{\tau^{*}}\xspace}
\newcommand{\optimalWaypointGt}{\ensuremath{q^{*}}\xspace}
\newcommand{\optimalWaypoint}{\ensuremath{\optimalWaypointGt}\xspace}
\newcommand{\optNetwork}{\ensuremath{f_{opt}}\xspace}

\newcommand{\snake}{8D-Snake\xspace}
\newcommand{\fetch}{11D-Fetch\xspace}

\newlength{\citeskipup}
\newlength{\citeskipdown}

\newboolean{bcmt}
\setboolean{bcmt}{true}
\usepackage{ifthen}

\usepackage{color}
\definecolor{darkgreen}{rgb}{0,0.5,0}
\definecolor{fullred}{rgb}{0.95,.0,.1}
\definecolor{orange}{rgb}{1,0.5,0}

\newcounter{ccmt}

\newcommand{\comment}[3]{\ifthenelse{\boolean{bcmt}}{\addtocounter{ccmt}{1}
  
  \marginpar{\tiny\noindent{\raggedright{{\colorbox{#3}{\sffamily\textcolor{white}{#1
            [\arabic{ccmt}]}}}}} \color{#3}{#2} \par}}{}}

\begin{document}

\title{Neural Randomized Planning for \\ Whole Body Robot Motion}





%


\author{
  Yunfan Lu\footnotemark[1] \qquad
  Yuchen Ma\footnotemark[1] \qquad
  David Hsu\footnotemark[1] \qquad
  Panpan Cai\footnotemark[2]\\
\normalsize
\parbox{3in}{
\begin{center} \textit{
  \footnotemark[1] Department of Computer Science\\
  National University of Singapore\\
  Singapore, 117417, Singapore\\}
\end{center}}
\parbox{3in}{
\begin{center} \textit{
  \footnotemark[2] Department of  Computer Science\\
  Shanghai Jiao Tong University\\
  Shanghai, 200240, China}
\end{center}}
}

\maketitle


\begin{abstract}
Robot motion planning has made vast advances over the past decades, but the challenge remains: robot mobile manipulators struggle to plan long-range
whole-body motion in common household environments in real time, because of high-dimensional robot configuration space and complex environment geometry.
To tackle the challenge, this paper proposes \emph{Neural Randomized Planner} (\NRP), which combines a global sampling-based motion planning (SBMP) algorithm and a local neural sampler. 
Intuitively, \NRP uses the search structure inside the global planner to ``stitch together” learned local sampling distributions to form a global sampling distribution adaptively. 
It benefits from both learning and planning. 
Locally, it tackles high dimensionality by learning to sample in promising regions from data, with a rich neural network representation. 
Globally, it composes the local sampling distributions through planning and exploits local geometric similarity to scale up to complex environments. 
Experiments both in simulation and on a real robot show \NRP yields superior performance compared to some of the best classical and learning-enhanced SBMP algorithms.
Further, despite being trained in simulation, \NRP demonstrates zero-shot transfer to a real robot operating in novel household environments, without any fine-tuning or manual adaptation.

\end{abstract}

\IEEEpeerreviewmaketitle


\section{Introduction}\label{sec:intro}

\begin{figure}[t]
    \centering
    \includegraphics[width=\columnwidth]{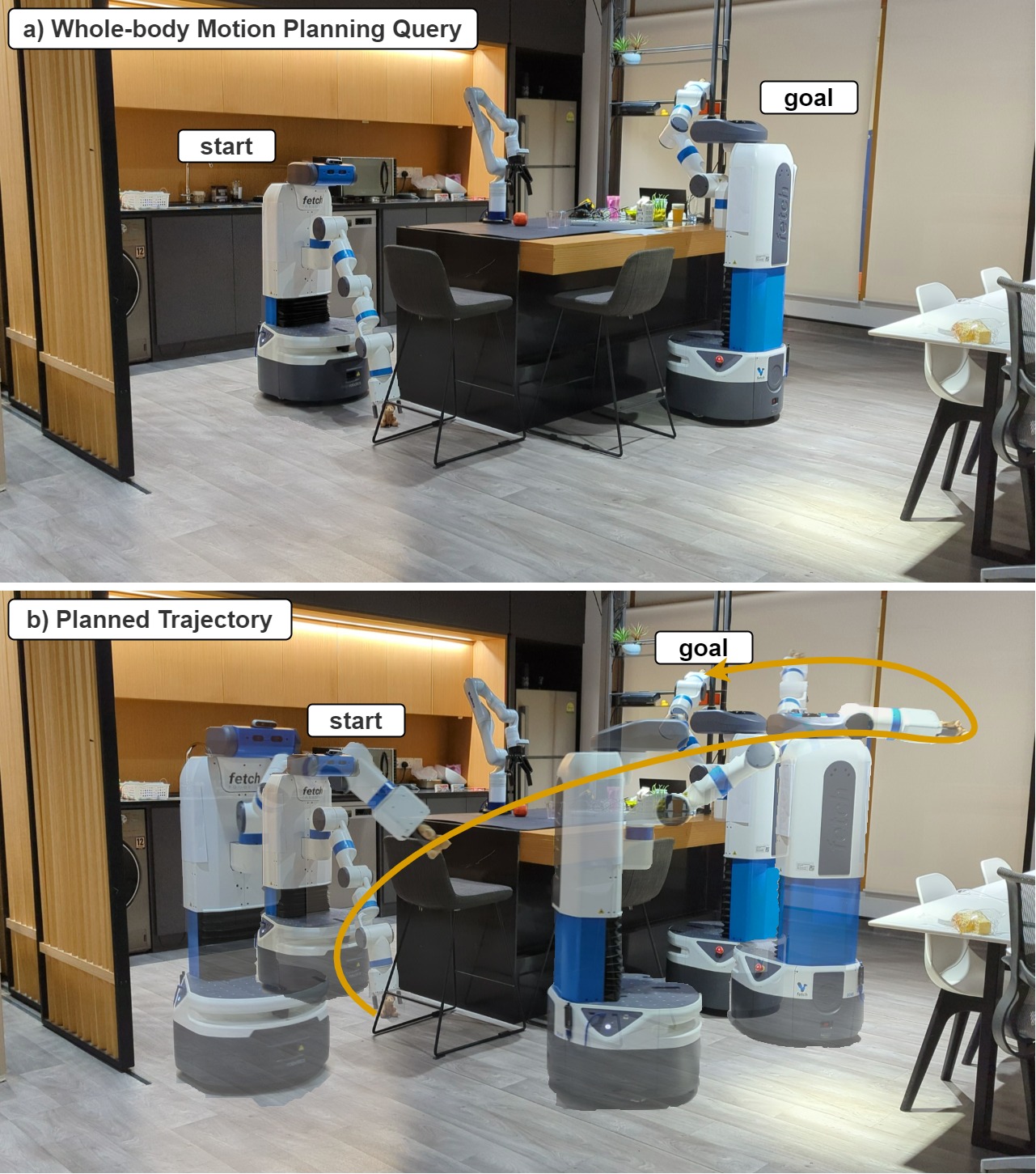}
    \caption{\textcolor{black}{A Fetch robot executing the whole-body motion planed by \NRP, with its end-effector trajectory shown in yellow. The task is to move a teddy bear on the ground to the kitchen shelf, requiring careful base and arm coordination for efficient obstacle avoidance and task accomplishment.}} \label{fig:wbmp}
    \vspace*{-10pt}
\end{figure}

Robot motion planning has made vast advances over the past
decades~\cite{elbanhawi2014sampling}\cite{mcmahon2022survey}. Yet, while humans maneuver their whole body with ease to accomplish complex long-range motions, robots struggle do the same in typical
household environments (\figref{fig:wbmp}). The key challenges are well-understood. 
First, robot mobile manipulators have many degrees of freedom (DoFs), leading to high-dimensional configuration spaces. Second, 
natural human environments, with doors, tables, chair, shelves, \etc,
add geometric complexity on top. Faced with these challenges,
sampling-based motion planing (SBMP) algorithms, which represent the state of the art, may take tens of seconds to process a motion planning query, and the computed motion is sometimes inefficient and awkward. To achieve human-level performance, we need optimized motion in a few seconds of computation time. How can we cross this gap? 

One most important lesson from the classic literature on motion planning is to craft an effective sampling distribution biased towards ``useful'' regions, such as collision-free spaces \cite{ichter2018learning, chamzas2019using}, narrow passages \cite{ichter2020learned, kumar2019lego}, or the neighborhoods  of optimal paths \cite{qureshi2019motion, jurgenson2019harnessing}.
Robot configurations sampled this way help SBMP algorithms to focus exploration on promising directions and thus find better solutions
faster. With the massive progress of deep learning in recent years, one natural extension is to learn the sampling distribution from data instead of crafting it manually. However, household environments have a huge variety of room and furniture designs, posing significant difficulties to learning sampling distributions that generalize over a wide range of household environments, with only limited data.  

To tackle this challenge, we observe that the variability in household and
many structured environments are \emph{compositional}. While different homes can have massively distinct layout, they all consist of kitchens, living rooms, bedrooms, \etc. Within each constituent space, the variability is much more contained. For example, two kitchens may be very similar, even if they are situated in very different homes. This suggests that we learn
\emph{local} sampling distributions and compose them into a global sampling
distribution, thereby exploiting the local geometric similarity to improve the data efficiency for learning.

Specifically, we propose \emph{Neural Randomized Planner}
(\NRP), with two main components: 
\begin{itemize}
\item \textbf{Global motion planner}.
Our sampling-based motion planner solves global whole-body motion planning queries by incrementally constructing and searching a planning tree over sampled configurations. It leverages samples from local sampling distributions learned offline to focus the search on promising directions.
\item \textbf{Local neural sampler}.
The neural sampler learns a distribution of waypoints that effectively
connects a configuration to a target configuration nearby, conditioned on a small local
environment. It encodes the sampling distribution in a neural network
representation. We explore two variants: a discriminative sampler and a
generative sampler. 
\end{itemize}
We first train the neural sampler using supervised learning, with data
collected from an expert motion planner. We then integrate the learned
sampler with a classic SBMP algorithm. There are many SBMP algorithms
available. For specificity, we consider two popular algorithms: RRT \cite{lavalle1998rapidly}
for feasible motion planning and Informed RRT* (IRRT*) \cite{gammell2014informed} for optimal
motion planning. \NRP uses the search tree in RRT or IRRT* to
``stitch together" the learned local sampling distributions and form a global sampling distribution adaptively.

We evaluated \NRP in the simulated Gibson \cite{xia2018gibson} environments and on the Fetch
robot. The results show that both \NRPRRT and \NRPIRRT outperform some of the
best classical and learning-based SBMP algorithms by a significant margin, in
terms of success rates and solution quality.
Further, while trained in simulation, \NRPRRT and
\NRPIRRT demonstrate zero-shot transfer to our Fetch robot operating in a
novel household environment, 
without any fine-tuning or manual  adaptation.

\textcolor{black}{
In summary, the core contribution of \NRP 
is a novel approach to integrating motion planning and learning. In the learning component, \NRP learns local sampling distributions for solving local planning problems optimally and adapts them to local environmental features using neural networks. This learning approach allows strong generalization in complex household environments. In the planning component, \NRP adaptively combines local sampling distributions through tree search to compute global whole-body motion. This systematic integration allows theoretical guarantees to be achieved in a straightforward way.
}

\begin{figure*}[ht]
    \centering
    \includegraphics[width=\textwidth]{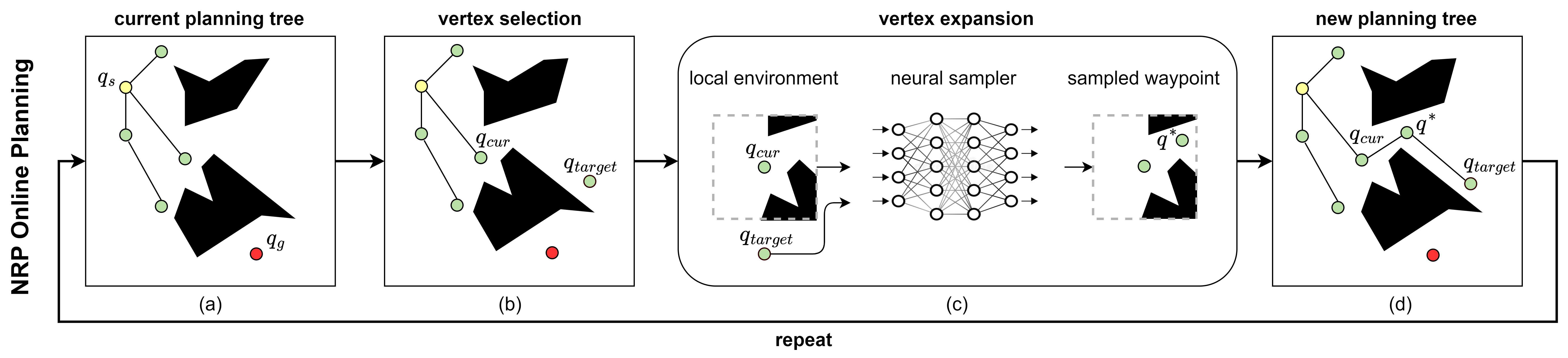}
    \caption{NRP Online Planning Process. (\subfig{a}) Initial planning tree with start configuration $\qstart$ and goal $\qgoal$. (\subfig{b}) An expansion target $\qtarget$ is sampled and the nearest vertex in the tree $\qcur$ is selected for expansion. (\subfig{c}) The neural sampler uses $\qcur$, $\qtarget$, and the local environment around $\qcur$ to generate an optimal waypoint $\optimalWaypoint$ for expansion. (\subfig{d}) The expansion path passing through $\optimalWaypoint$, upon passing collision checks, is added to the tree. This sequence repeats until a solution is discovered or a maximum planning time is reached.}
    \vspace*{-10pt}
    \label{fig:online_planning_overview}
\end{figure*}

\section{Related work}

\subsection{Generating Whole-Body Robot Motions}
Sampling-based motion planning (SBMP) methods \cite{luna2020scalable, burget2013whole, dai2014whole, stilman2010global, wei2019motion, haddad2006optimal}represent the state of the art of planning long-range whole-body motions in large environments. They use random samples to construct a discrete approximation of the robot configuration space and perform graph searches to compute feasible or optimal motion plans. SBMP methods guarantee probabilistic completeness and asymptotical optimality of solutions. 
However, the core challenges lie in tackling narrow passages and low-dimensional tunnels, which are difficult to cover with random sampling. These challenges are severe when considering high-DOF mobile manipulators and complex household environments, where classical SBMP algorithms become too inefficient for practical use. 
Existing methods aimed to accelerate planning either decomposed the arm and base motions \cite{stilman2007manipulation, castaman2021receding, saoji2020flexibly, rastegarpanah2021semi}, actuating either arm or base at each particular path segment, or separately planned the arm and base motion, one conditioned on the other \cite{chitta2010planning, chitta2012mobile}.
However, these simplifications sacrificed the completeness and optimality guarantee of planning, leading to high failure rates and lower-quality paths.

\textcolor{black}{Recently, learning approaches, such as reinforcement learning \cite{jurgenson2019harnessing, strudel2021learning} and imitation learning \cite{qureshi2019motion, kicki2021learning, kicki2022speeding, kicki2024fast, qureshi2020neural, li2021mpc, qureshi2022constrained}, have gained increased popularity in generating robot motions. However,
these works mainly focused on low-dimensional planning in small workspaces, \eg, tabletop environments.
Directly learning an end-to-end policy for high-dimensional problems has been proven ineffective \cite{xia2021relmogen, hu2023causal}. 
To scale up, some works explored similar simplifications as in SBMP to assist learning.}
For instance, \cite{xia2021relmogen, li2020hrl4in} learned a decomposed control policy, which only actuated either arm or base at every time step. 
\cite{jauhri2022robot, honerkamp2023n} separately learned two policies for arm and base, one conditioned on the other.
These simplifications often led to low success rates and inefficient motion plans due to the lack of whole-body coordination. 
Other learning approaches leveraged a factored reward function \cite{hu2023causal} or a carefully designed curriculum \cite{fu2023deep} to improve sample efficiency. 
However, their application was mostly limited to reaching nearby goals among a few obstacles, lacking demonstration of long-term motion planning capabilities in complex household environments. 


\subsection{Learning Sampling Distributions}

A key idea to achieve efficient global planning while keeping desirable theoretical guarantees is to bias the sampling distribution for SBMP algorithms, directing the search to valuable regions of the configuration space \cite{urmson2003approaches, hsu2006probabilistic}, thus accelerating planning. With the massive progress of deep learning in recent years, learning such a sampling distribution from data has become a popular choice. Previous methods have mainly taken three approaches: 1) \textcolor{black}{learning to sample in the collision-free  
\cite{tran2020predicting, kew2019neural, das2020learning, yu2021reducing, chamzas2019using, chamzas2021learning} or kinodynamically-feasible \cite{lembono2021learning, wolfslag2018rrt, atreya2022state, yavari2019lazy} space, through generative learning or combining a learned discriminator with rejection sampling}; 2) learning to sample in critical regions (e.g., near narrow passages) from supervised data \textcolor{black}{\cite{kumar2019lego, khan2020graph, molina2020learn, ichter2020learned}}; and 3) learning to sample around optimal paths to the global planning problem, \textcolor{black} {using supervised learning \cite{ichter2020learned, chen2019learning, schramm2022learning, huh2018efficient, cheng2020learning, kobashi2023learning} or reinforcement learning \cite{zhang2018learning}}. We refer to these distributions as collision-free, critical, and optimal sampling distributions.
However, directly learning a single global distribution over global environments makes it difficult for these methods to generalize over long-range tasks in complex household environments.
Therefore, \cite{chamzas2019using, chamzas2021learning, chamzas2022learning} learned a global sampling distribution by explicitly stitching elementary sampling distributions, each for collision-free or critical sampling regarding a local feature of the environment. 
Generalization is achieved by matching new features to existing ones stored in a database.
This approach was evaluated in tasks that have small workspaces, such as tabletop manipulation, where a limited number of local features, like objects on the table, could be manually defined.
\textcolor{black}{However, in large household environments with an enormous amount of local features, the size required for the database to adequately represent possible local features grows exponentially, impacting the method's performance.} 
Furthermore, compared to optimal sampling, critical samples only offer limited benefits to computing the optimal solution. 
Different from the above methods, \NRP learns local sampling distributions for optimal sampling using a neural network, and leverages the search structure in SBMP algorithms to adaptively compose the learned local distributions into a global one.
\NRP enjoys greater data efficiency compared to directly learning global sampling distributions and better generalization compared to explicit feature matching.

\section{Neural Randomized Planning}
We propose Neural Randomized Planner (\NRP), a motion planning algorithm for high-DOF robots in complex household environments. \NRP comprises a global planner and a local planner. The global planner constructs and searches a tree across random global configurations to reach the goal, while the local planner connects the tree to nearby global samples using a neural sampler. 
The neural sampler is trained offline in simulation, supervised by an expert SBMP algorithm.

\subsection{Global Motion Planner}



The \NRP's global motion planner leverages the local neural sampler to focus exploration on promising directions. It is built upon two established SMBP algorithms: Rapidly-exploring Random Tree (RRT) \cite{lavalle1998rapidly} and Informed-RRT* (IRRT*) \cite{gammell2014informed}, facilitating probabilistically complete and asymptotically optimal planning. 
We name the resultant algorithms \NRPRRT and \NRPIRRT. 

\figref{fig:online_planning_overview} presents an overview of the proposed
planners. \NRPRRT and \NRPIRRT incrementally expand a planning tree to connect
with random configurations sampled in the entire configuration space (C-space), thus, eventually, the goal. 
At every planning iteration, \NRPRRT samples a configuration uniformly and treats it as the tree expansion target $\qtarget$. Then, it selects a vertex $\qcur$ in the current planning tree nearest to $\qtarget$ for expansion (\textit{vertex selection}). 
In order to connect $\qcur$ and $\qtarget$ (\textit{vertex expansion}), \NRPRRT inputs $\qcur$, $\qtarget$, and the local environment around $\qcur$ into the neural sampler to obtain an \emph{optimal waypoint}, denoted as $\optimalWaypoint$. Formally, $\optimalWaypoint$ is defined as a configuration that lies along the optimal path $\solPathOptimal$ from $\qcur$ to $\qtarget$, resides within the local environment surrounding $\qcur$ and is connectable to $\qcur$ via C-space shortest paths (\eg, straight-lines in Euclidean spaces).
A vertex expansion path is then constructed by connecting $\qcur$ through $\optimalWaypoint$ to $\qtarget$, and the collision-free portion is added to the planning tree. While \NRP proceeds to the next iteration, \NRPIRRT performs shortcut and rewiring operations after each expansion to achieve optimality, similar to IRRT*.
Additionally, with a small probability, the algorithm reverts to the shortest path expansion in native RRT and IRRT*. This retains the probabilistic completeness and asymptotic optimality guarantee of the original algorithms.

\subsection{Local Neural Sampler}

\begin{figure*}[t]
    \centering
    \includegraphics[width=\textwidth]{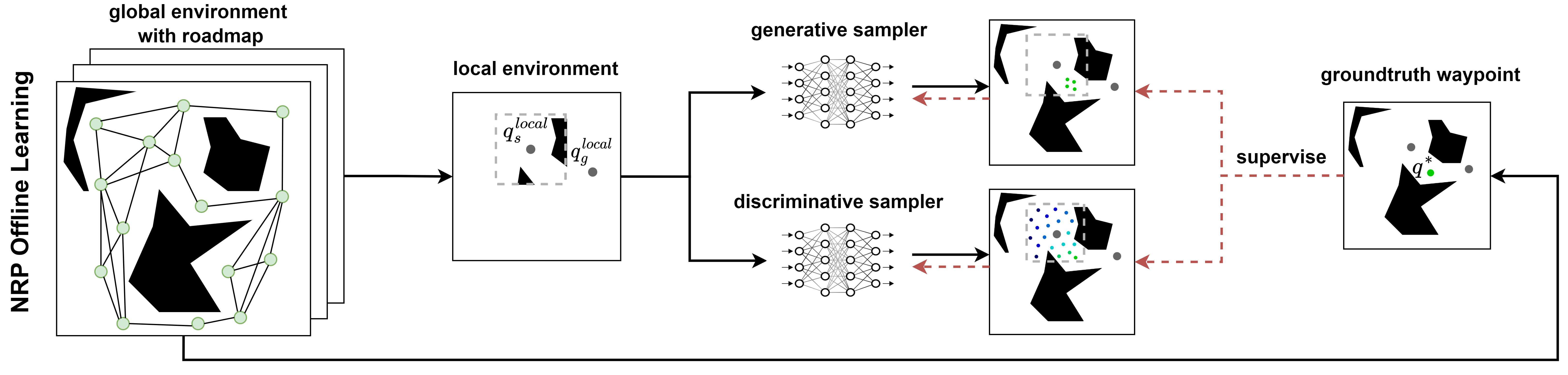}
    \caption{\NRP's Offline Learning Process. Training data is collected by sampling numerous planning queries, \ie, start and goal configurations, within training environments. For each query, the ground truth optimal waypoint, denoted as $\optimalWaypoint$, is extracted from the roadmap pre-computed by an expert motion planner. This waypoint serves as supervision for both local neural samplers. Specifically, the generative sampler learns to reconstruct optimal waypoint from latent-space samples, while the discriminative sampler learns to classify optimal waypoints from random samples. }
    \label{fig:offline_learning_overview}
    \vspace{-10pt}
\end{figure*}

The \NRP's local neural sampler (\figref{fig:offline_learning_overview}) generates optimal waypoints $\optimalWaypoint$ for efficiently connecting nearby configurations, represented by the local start $\qLocalStart$ and the local goal $\qLocalGoal$, conditioned on the local environment $\Wlocal$ surrounding $\qLocalStart$.
We develop two versions of the neural sampler: a discriminative one, which implicitly encodes the distribution of $\optimalWaypoint$ using neural network discriminators; and a generative one, which explicitly samples from the distribution using a latent-space model.
Both samplers are trained using supervised learning, with data from an expert SBMP algorithm, Probabilistic Roadmap* (PRM*) \cite{karaman2011sampling}.

\subsubsection{Discriminative Sampler}
\begin{figure}[t]
    \centering
    \includegraphics[width=\columnwidth]{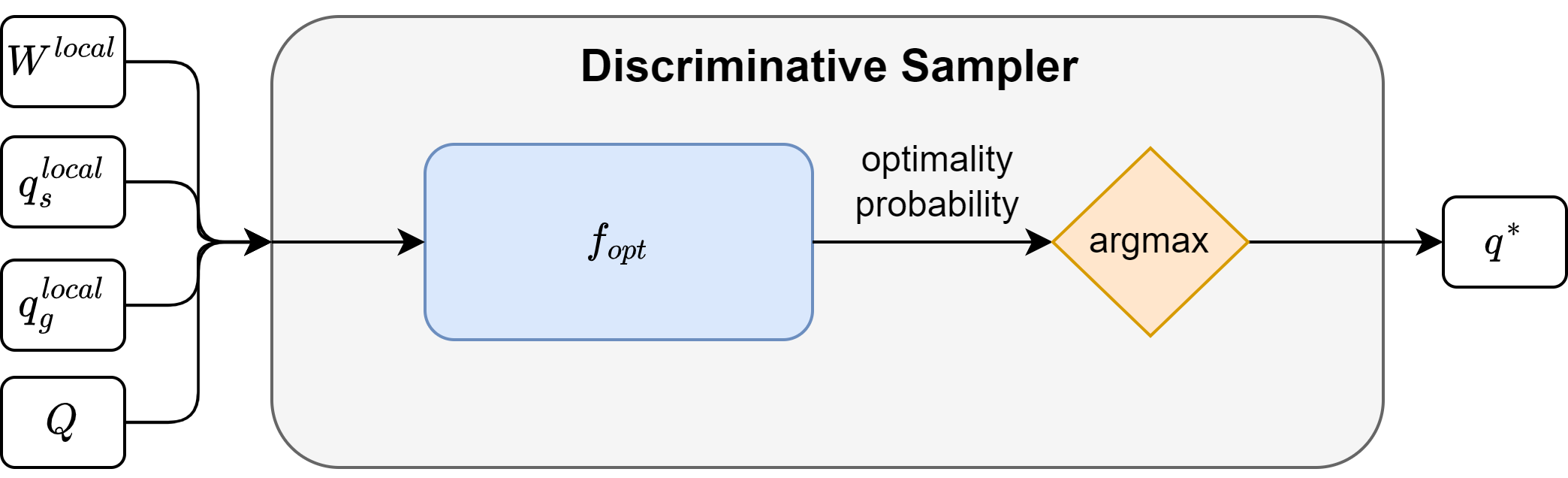} 
    \caption{\textcolor{black}{Overview of discriminative sampler. Q denotes a set of uniformly sampled  configurations within $\Wlocal$.}}
    \label{fig:learner_architecture_d}
    \vspace{-10pt}
\end{figure}

\textcolor{black}{The discriminative sampler (\figref{fig:learner_architecture_d}) encodes the distribution of optimal waypoints $p(\optimalWaypoint|\qLocalStart, \qLocalGoal, \Wlocal)$ using an optimality discriminator $\optNetwork$, which assesses whether $q$ lies on the optimal path to the local goal, $\qLocalGoal$:}

\begin{equation}
    p_{q}^{opt} = \optNetwork(\qLocalStart, \qLocalGoal, \Wlocal, q)
\end{equation}

\textcolor{black}{$\optNetwork$ is a feedforward neural network, using 3D Convolution layers to extract features from the occupancy grid of the local environment. Full details of the input and network architecture are included in Appendix C. $\optNetwork$ is trained using Binary Cross-Entropy (BCE) loss, using ground truth optimality labels $ y_{q}^{opt}$ in the dataset:
\begin{equation}
    L_{opt} = -\frac{1}{N} \sum_{i}^{N} y_{q_{i}}^{opt} \log(p_{q_{i}}^{opt}) + (1 - y_{q_{i}}^{opt}) \log(1 - p_{q_{i}}^{opt}) 
\end{equation}
}
\textcolor{black}{
To obtain a sample from the discriminative sampler, a set of candidate configurations are first uniformly sampled within $\Wlocal$, then the optimal waypoint $\optimalWaypoint$ is determined as the one with the highest predicted probability $p_{q}^{opt}$.
}
\textcolor{black}{
\begin{equation}
    \optimalWaypoint = \argmax_{q \sim Uniform}p_{q}^{opt}
\end{equation}
}

\subsubsection{Generative Sampler}

\begin{figure}[t]
    \centering
    \includegraphics[width=\columnwidth]{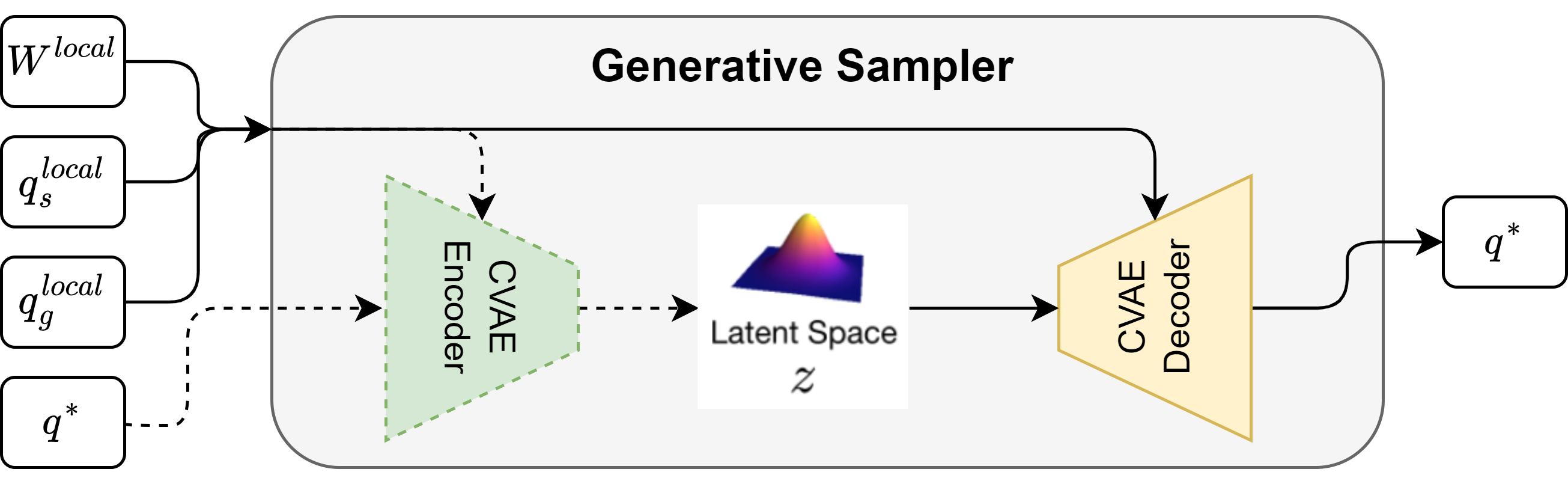} 
    \caption{Overview of generative sampler. \optimalWaypoint denotes the optimal waypoint. We only use the decoder at inference time.
    }
    \label{fig:learner_architecture_g}
    \vspace{-10pt}
\end{figure}

The generative sampler (\figref{fig:learner_architecture_g}) utilizes a Conditional Variational AutoEncoder (CVAE) \cite{sohn2015learning} to explicitly sample from the distribution of optimal waypoints $p(\optimalWaypoint|\qLocalStart, \qLocalGoal, \Wlocal)$, with the help of a latent variable $z$.
The CVAE architecture consists of an encoder and a decoder. The encoder, parameterized by $\phi$, models the conditional probability $p_{\phi}(z|\optimalWaypoint, \qLocalStart, \qLocalGoal, \Wlocal)$, outputting the mean and variance of the latent variable $ z $ based on the context $ C $ (comprising $\Wlocal$, $\qLocalStart$, and $\qLocalGoal$) and the waypoint $\optimalWaypoint$. The decoder, parameterized by $\theta$, then reconstructs the optimal waypoint from $z$, modeling $p_{\theta}(\optimalWaypoint|z, C)$.

CVAE is trained by maximizing an Evidence Lower Bound (ELBO), using ground-truth samples of $\optimalWaypointGt$:
\begin{equation} \label{eq:ELBO}
\begin{split}
ELBO = & E_{z \sim q_{\phi}(z|\optimalWaypointGt,C)}[\log p_{\theta}(\optimalWaypointGt|z,C)]- \\
       & D_{KL}(q_{\phi}(z|\optimalWaypointGt,C) \; || \; \mathcal{N}(\mathbf{0}, I)),
\end{split}
\end{equation}
where $D_{KL}$ represents the Kullback-Leibler divergence. This objective ensures that the decoder is trained to accurately reconstruct optimal waypoints while keeping the encoder's latent space distribution close to a standard normal distribution. \textcolor{black}{The network architecture of the generative sampler is very similar to the discriminative one. See details in Appendix C.}

During planning, the generative sampler explicitly samples a value of $z$ from a standard normal distribution and inputs it, along with the context $C$, into the decoder which subsequently generates an optimal waypoint $\optimalWaypoint$.

\subsubsection{Training Dataset}

The training dataset for samplers is generated with the help of an expert sampling-based motion planner, Probabilistic Roadmap* (PRM*), which is used to construct a dense roadmap in each training environment $\Wglobal$ at the start of data collection. 
Afterwards, a set of planning queries is sampled from each environment $\Wglobal$, denoted as tuples containing local start $\qLocalStart$, local goal $\qLocalGoal$, and local environment $\Wlocal$ around the start. 
For each tuple, the optimal path $\solPathOptimal$ from $\qLocalStart$ to $\qLocalGoal$ is computed using the constructed roadmap, considering only obstacles within \Wlocal to maintain the local context of the planning problem.
The optimal waypoint for each query is identified as the first waypoint on $\solPathOptimal$ that lies within $\Wlocal$. This waypoint serves as the ground truth $\optimalWaypointGt$ for training the samplers.

To augment the training dataset for the discriminative sampler, additional waypoints are randomly sampled within $\Wlocal$.\
\textcolor{black}{
Each sampled waypoint is then given an optimality label $y_{q_{i}}^{opt}$ , determined by whether the optimal path length from $\qLocalStart$ to $\qLocalGoal$, via the waypoint, approximates path length of $\solPathOptimal$. 
} 
\textcolor{black}{
This provide the discriminative sampler with a balanced dataset of optimal and sub-optimal waypoints, enhancing its ability to discern effective samples.
}

\begin{figure*}[t]
    \centering
    \subfloat[]{\includegraphics[width=0.49\textwidth]{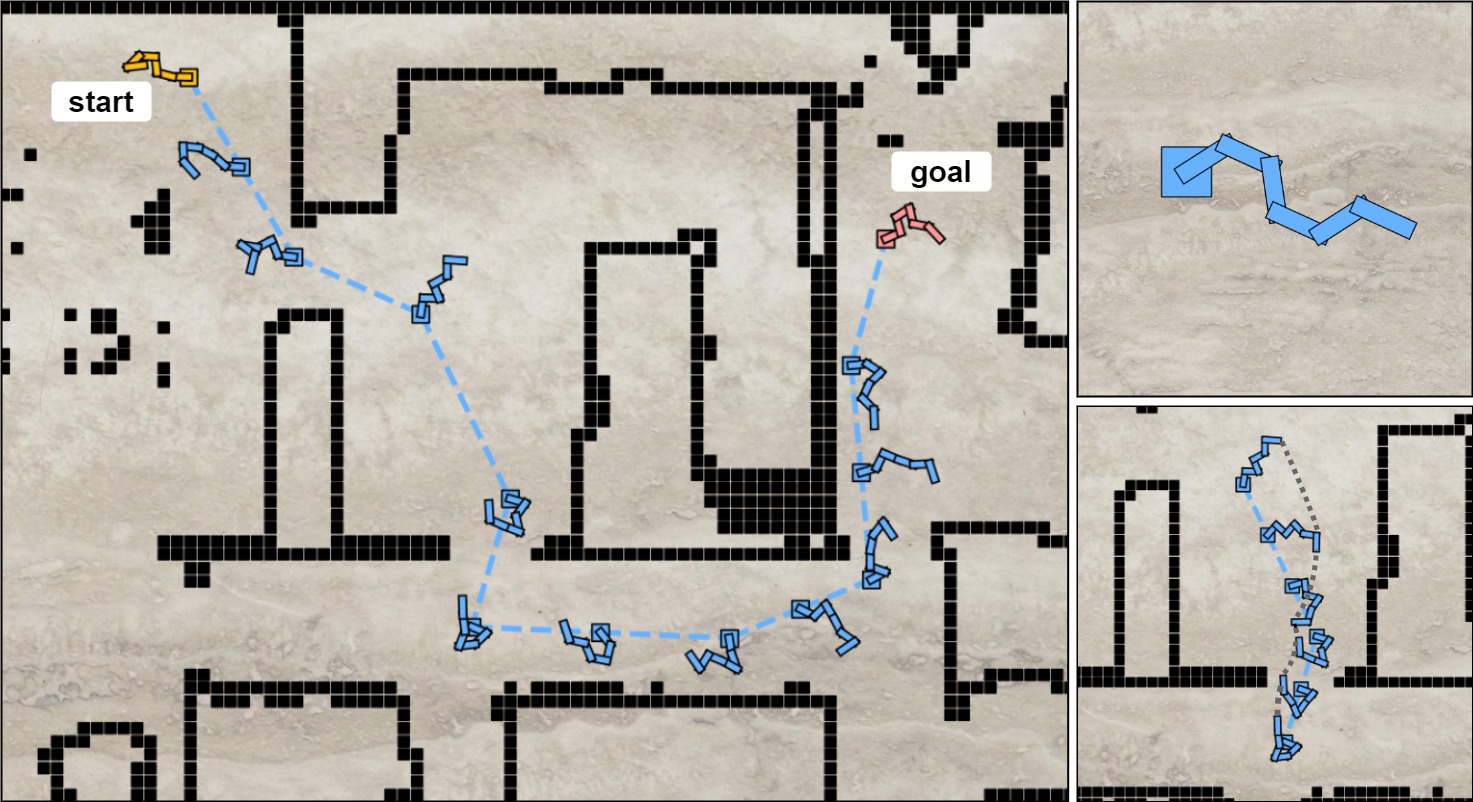}}\hfill
    \subfloat[]{\includegraphics[width=0.49\textwidth]
    {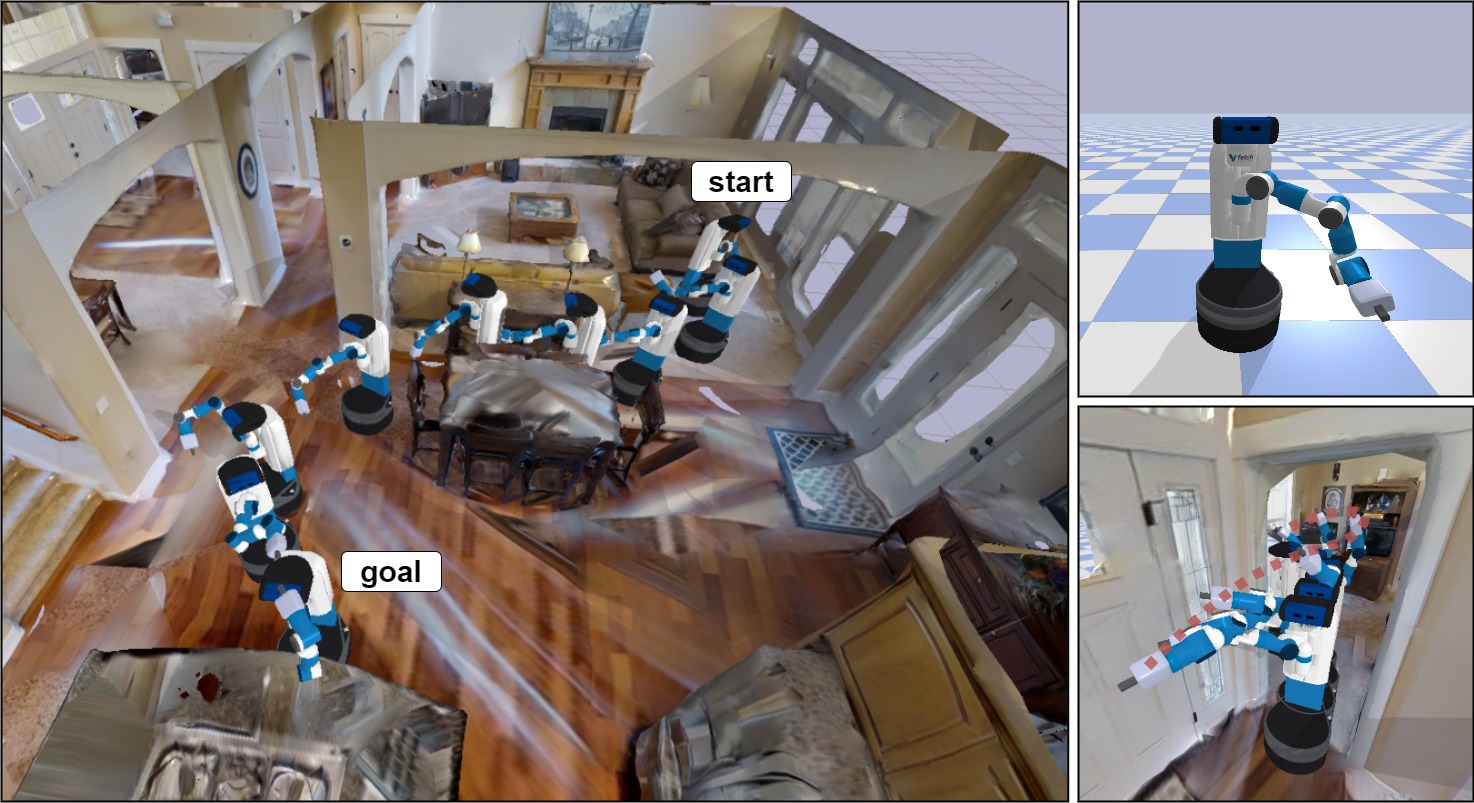}}\hfill
    \caption{Visualization of motion planning tasks in simulation. (\subfig{a}) Visualization for \snake task. (\subfig{b}) Visualization for \fetch task. In each case, the top-right figure offers a detailed perspective of the robots, while the bottom-right figure emphasizes the need for careful coordination between the robot's arm and base to navigate through narrow corridors in the test environments.}
    \label{fig:env_viz}
    \vspace{-10pt}
\end{figure*}

\section{Simulation Experiments}


This section presents simulation experiments where \NRPRRT and \NRPIRRT are evaluated in complex motion planning tasks in realistic household environments, and compared with state-of-the-art classical SBMP algorithms and their variants incorporating learned sampling distributions.
In simulations, \NRP demonstrates substantial advantages in success rates and path optimality across different robots and various household environments. The findings highlight \NRP's robustness and generalizability to novel environments. 
Deeper analysis revealed that \NRP's superior performance arises from high-quality sampling distributions learned by its local neural samplers, which increase the exploration efficiency of planning and generalization across environments. 


\subsection{Experimental Setup in Simulation}
Our simulation tasks closely mimic real-world long-range motion challenges within household settings, constructed using environments from the Gibson dataset \cite{xia2018gibson}.

\subsubsection{Evaluation Tasks}

The first task, referred to as \snake, involves moving a snake robot through 2D maps of real houses. This robot features a square-shaped freely moving base and a 6-DOF arm, resulting in a total of 8 DOFs. The second task, \fetch, involves navigating a Fetch robot in 3D houses. The Fetch robot, equipped with a mobile base, a controllable torso, and a 7-DOF arm, has a total of 11 DOFs. Both tasks demand strategic arm positioning to navigate through tight corridors and doorways, depicted in \figref{fig:env_viz}.

 \subsubsection{Variants and Comparison Baselines}

In our experiments, we evaluate the following variants of \NRP:
\begin{itemize}
    \item \textbf{\NRPRRT-d} and \textbf{\NRPIRRT-d}: Variants of \NRP employing the discriminative sampler for feasible and optimal planning, respectively.
    \item \textbf{\NRPRRT-g} and \textbf{\NRPIRRT-g}: Variants of \NRP that utilize the generative sampler, again divided into feasible and optimal planners.
\end{itemize}

These \NRP configurations are compared against a suite of established SBMP algorithms, including traditional approaches:
\begin{itemize}
    \item \textbf{RRT}: The original Rapidly-exploring Random Tree algorithm.
    \item \textbf{RRT-IS}: An RRT variant that adds intermediate configurations along the vertex expansion paths to the planning tree.
    \item \textbf{IRRT*}: The original Informed RRT*, an algorithm for asymptotically optimal planning.
    \item \textbf{BIT*} \cite{gammell2015batch}: The Batch Informed Trees algorithm, known for efficient, optimal planning.
\end{itemize}
and SBMP algorithms with learned sampling distributions:
\begin{itemize}
    \item \textbf{CVAE-RRT} \cite{ichter2018learning}: Combines RRT with a global sampling distribution learned using a Conditional Variational AutoEncoder.
    \item \textbf{CVAE-IRRT*} \cite{ichter2018learning, wang2020neural}: Similar to CVAE-RRT but employs IRRT* for planning.
    \item \textbf{NEXT} \cite{chen2019learning}: Employs a policy network for tree expansion towards the goal, supported by a value network for vertex selection, both conditioned on the global environment.
    \item \textcolor{black}{\textbf{\FIRE} \cite{chamzas2022learning}: Combines RRT with a global sampling distribution, which is constructed by retrieving relevant local sampling distributions from a database, based on a learned similarity function.}
    \item \textcolor{black}{\textbf{\FIREStar} \cite{chamzas2022learning}: Similar to \FIRE but employs IRRT* for planning.}
\end{itemize}

\subsubsection{Implementation Details}

\NRP was trained separately for each task. 
The training spanned across 25 diverse environments, from which a dataset of 100,000 samples was derived. Each environment contributed 500 queries, from which we collected 8 waypoints per query.
In the case of the discriminative sampler, these waypoints included both optimal and sub-optimal instances. Conversely, for the generative sampler, all collected waypoints were optimal.
We trained Learning-based baselines using the same amount of data to ensure a fair comparison. 
For CVAE-RRT and CVAE-IRRT*, the same 500 queries were sampled in each training environment. For each query, 8 waypoints were uniformly discretized from the optimal path to form the training dataset.
NEXT adheres to its original, online training protocol \cite{chen2019learning}, where the planner itself continuously generates new data for the learner, by using the latest policy and value networks for planning. Training continues until encountering the same total amount of data as \NRP. \textcolor{black}{For \FIRE, the same 500 queries were used to learn local environmental features and the corresponding local distributions, which formed the database.} 

Testing used 5 unseen environments, where 50 queries were sampled per environment to form a set of 250 test queries.
\textcolor{black}{We compare the performance of algorithms when given the same amount of planning time, ranging from 0 up to 10s. 
Specifically, the planning time accounts for all online computations, including neural network inference. However, it does not account for the pre-processing time of constructing the occupancy grid for the global environment. Note that the occupancy grid is supplied to all planning methods to expedite online collision checking. NRP does not incur additional time for collision reasoning compared to other methods.
}

All reported results are the averages across these queries, consolidated over 10 independent runs to ensure reliability and consistency, \textcolor{black}{with standard deviations represented by the shaded region.} 

In the \snake task, 
\NRP defined the local environment as a $4m \times 4m$ square centered around the robot's current location. 
For the \fetch task, \textcolor{black}{a local environment becomes a $4m \times 4m \times 2m$ box}. All environments were represented by an occupancy grid with a resolution of $0.1m$. To account for the non-holonomic nature of the Fetch robot's base, Reeds-Shepp curves \cite{reeds1990optimal} were employed as the shortest paths in the configuration space.

\textcolor{black}{Note that \FIRE was evaluated only on \fetch task as it has been designed for 3D environments. The original local environment dimensions of $0.4m \times 0.4m \times 0.4m$ resulted in a database size of approximately 5,000,000 for our global environments, 500 times larger than the database size in \cite{chamzas2022learning}. Consequently, it took more than 10 minutes to construct a global sampling distribution for each query, rendering it impractical. Instead, we set the local environment size to be $4m \times 4m \times 2m$, same as \NRP, to reduce the construction time to around 13s. This time was excluded in the planning time for \FIRE and \FIREStar.
}

All algorithms were implemented in Python, following the methodologies presented in their respective publications, with the code optimized to maximize performance. The experiments were conducted on the same PC with Intel Xeon CPU @ 3.00Ghz and a Nvidia 2080 Ti GPU.

\subsection{Performance Comparison with Existing Approaches}

\begin{figure}[t]
    \centering
    \includegraphics[width=\columnwidth]{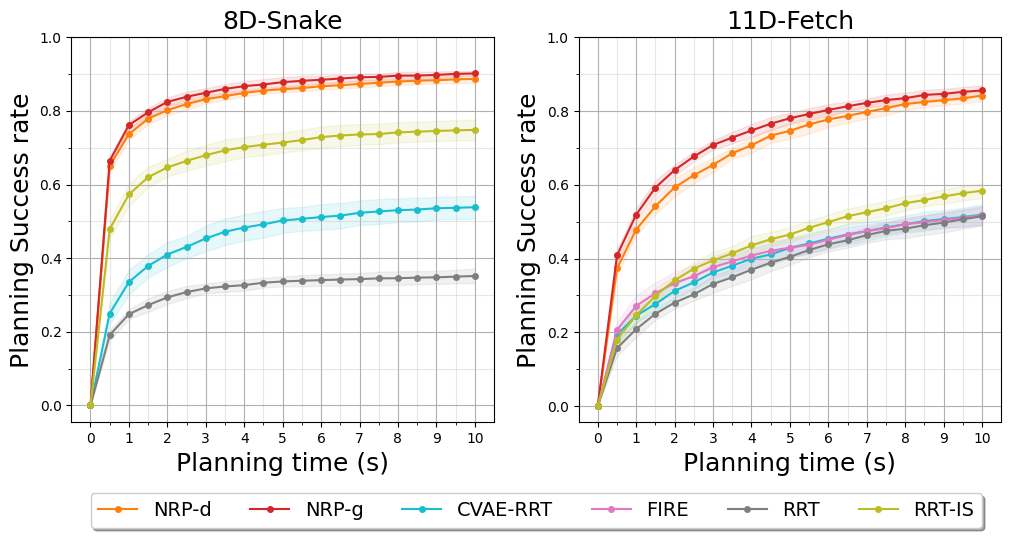}
    \caption{\textcolor{black}{Average planning success rate for feasible planners.}}
    \label{fig:planning_res}
    \vspace{-10pt}
\end{figure}

\begin{figure}[t]
    \centering
    \includegraphics[width=\columnwidth]{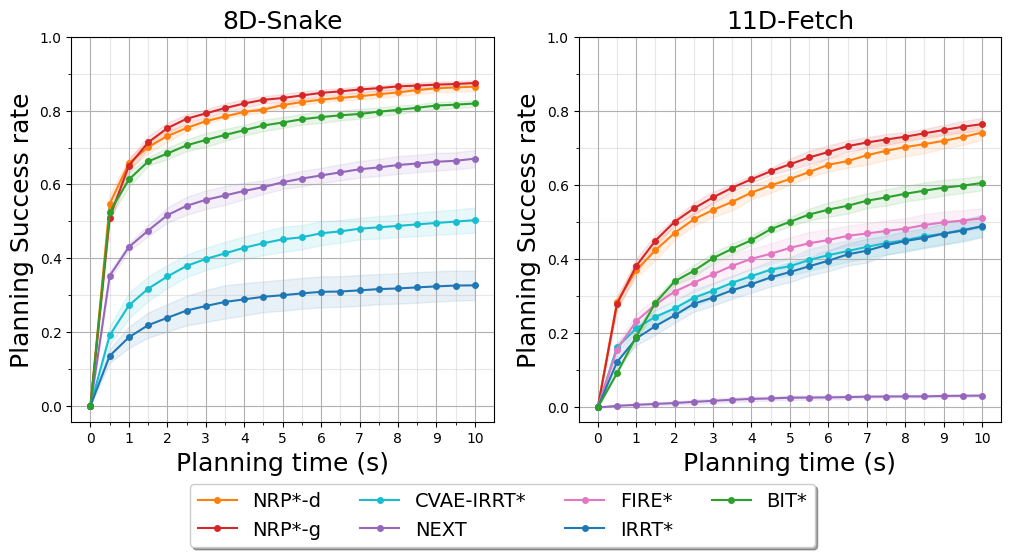}
    \caption{\textcolor{black}{Average planning success rate for optimal planners.}}
    \label{fig:planning_res_optimal}
    \vspace{-10pt}
\end{figure}

\begin{figure}[t]
    \centering
    \includegraphics[width=\columnwidth]{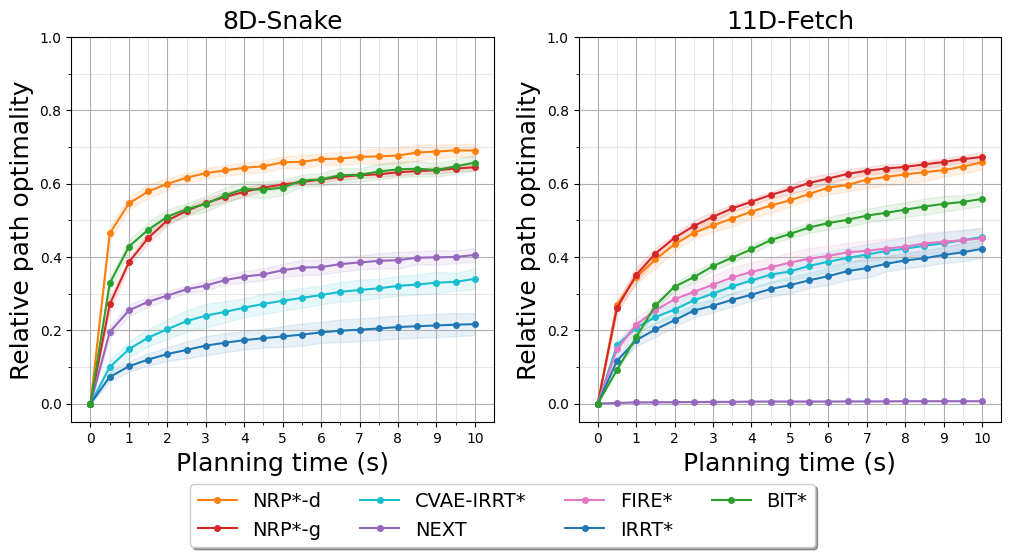}
    \caption{\textcolor{black}{Average relative path optimality for optimal planners.}}
    \label{fig:path_cost_optimal}
    \vspace{-10pt}
\end{figure}

The comparative analyses, illustrated in \figref{fig:planning_res}, \ref{fig:planning_res_optimal} and \ref{fig:path_cost_optimal}, demonstrate that \NRP consistently achieved high success rates and path optimality across the challenging \snake and \fetch tasks, outperforming classical SBMP algorithms such as RRT, RRT-IS, IRRT*, and BIT*, as well as advanced planners like CVAE-RRT, CVAE-IRRT*, NEXT, \textcolor{black}{\FIRE} and \textcolor{black}{\FIREStar}

In the context of feasible planning (\figref{fig:planning_res}), \NRPRRT achieved around \textcolor{black}{20\%} higher success rates compared to RRT-IS, the top-performing feasible SBMP algorithm, and showed a significant \textcolor{black}{64\%} improvement over CVAE-RRT, the leading planner among those with learned sampling distributions, in the \snake task.
This advantage became more significant in the \fetch task.
\NRPRRT consistently attained success rates approximately \textcolor{black}{42\% higher than RRT-IS and 70\% higher than CVAE-RRT}, showcasing the advantage of \NRPRRT in higher-dimensional and more intricate environments. 

In the context of optimal planning (\figref{fig:planning_res_optimal}), \NRPIRRT surpassed NEXT, the best optimal planner with learned sampling, by about \textcolor{black}{30\%} and achieving a slight edge over the classical optimal planning algorithm BIT* by \textcolor{black}{7\%} in the \snake task. The performance gap became more pronounced in the \fetch task, with \NRPIRRT \textcolor{black} {achieving around 50\% more success than {\FIREStar} (even though the latter additionally used 13 seconds for distribution construction) and 25\% more success than BIT*}.
The evaluation of path optimality (\figref{fig:path_cost_optimal}) further highlights \NRPIRRT's efficiency in optimal planning. Path optimality is measured by comparing the length of the optimal path to that of the computed path, averaged over all queries. Results in \figref{fig:path_cost_optimal} demonstrate that \NRPIRRT finds \textcolor{black}{higher quality} solutions more swiftly, particularly in the demanding \fetch task. 
The superior performance of \NRP highlights the effectiveness of learned sampling distributions. 

\begin{table*}[t]
\caption{Comparison of the exploration and exploitation efficiency of RRT and \NRP variants in \snake}
\label{table:vertex_exp_quality}
\begin{footnotesize}
\centering
\begin{tabularx}{1\textwidth}{l@{\extracolsep{\fill}}cccc}
    \toprule
    \multirow{3}{*}{} Algorithm &  Average time & Average time & Average expansion & Average expansions \\
     & needed per unit of & needed per unit of & needed per unit of & needed per unit of \\
    & visibility growth (ms) & dist-to-goal shortened (ms) & visibility growth & dist-to-goal shortened \\
    \midrule
    RRT & 1.70 & 266 & 1.742 & 208.9   \\
    \NRPRRT-d & \textbf{0.70} & 123 & \textbf{0.227} & 35.51 \\
    \NRPRRT-g & 0.86 & \textbf{111} & 0.308 & \textbf{30.99} \\
    \bottomrule
\end{tabularx}
\end{footnotesize}
\end{table*}

\begin{figure}[t]
    \centering
    \hfill
    \subfloat[]{\includegraphics[width=0.49\columnwidth]{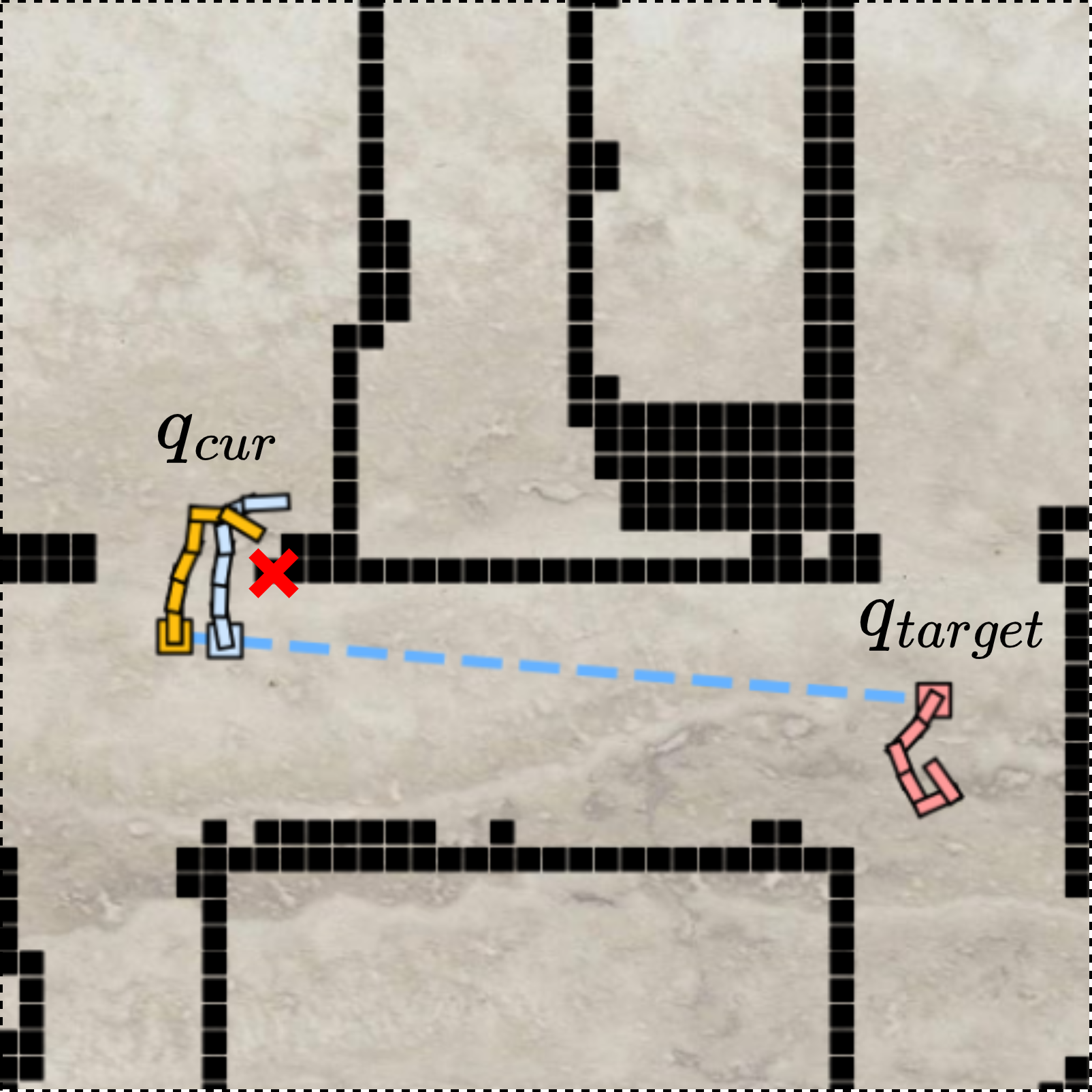}}\hfill
    \subfloat[]{\includegraphics[width=0.49\columnwidth]{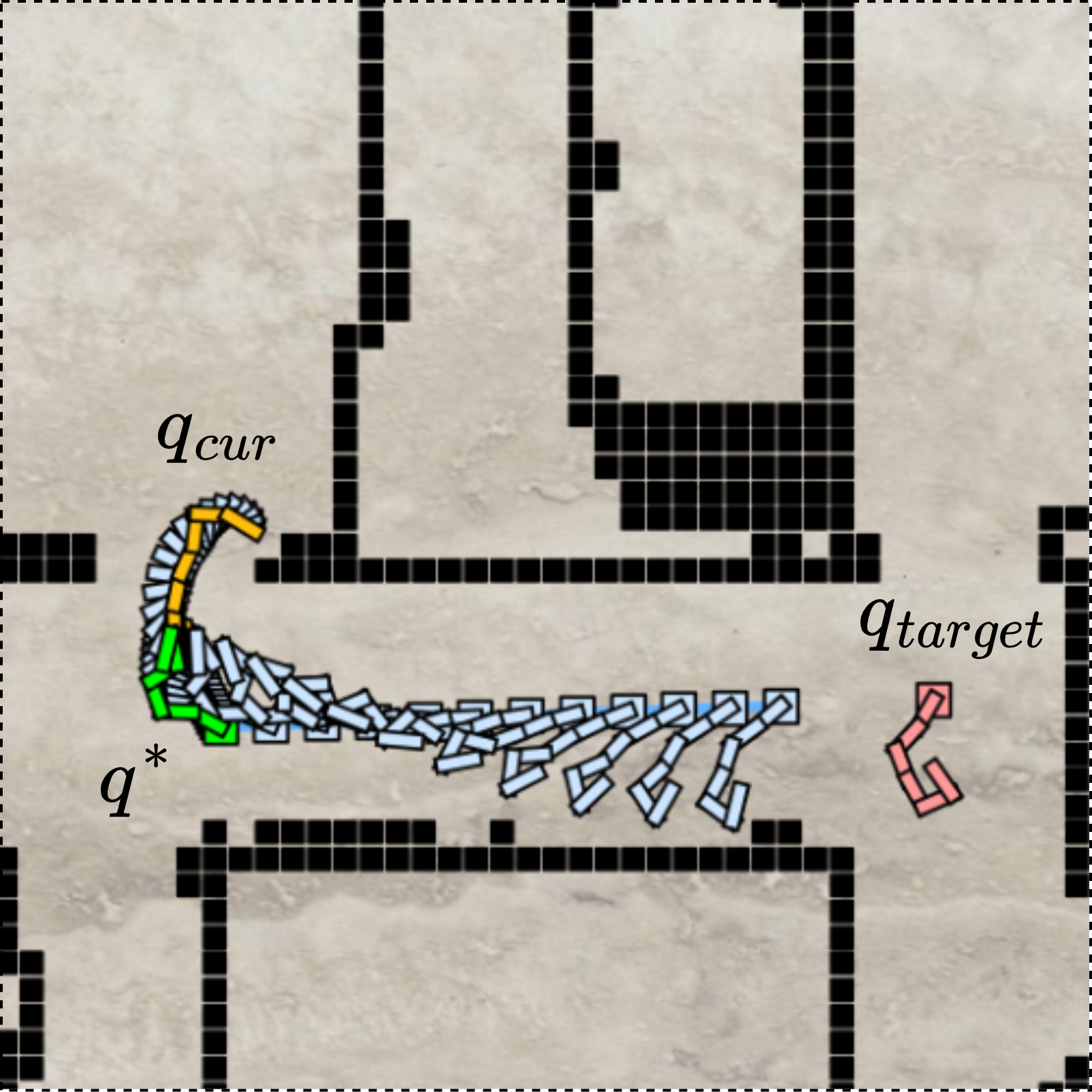}}\hfill
    \caption{\textcolor{black}{Visualization of vertex expansion path in the \snake task. 
    (\subfig{a}) A straight-line expansion in RRT leads directly to a collision. (\subfig{b}) \NRP's neural sampler produced a waypoint $\optimalWaypoint$ (green) that effectively coordinates the base and arm motion, enabling a closer approach to $\qtarget$.
    }}
    \label{fig:viz_ext_8d}
    \vspace{-10pt}
\end{figure}

\subsection{The Benefit of Learning}

Further analysis in \tabref{table:vertex_exp_quality} demonstrates that learned sampling distributions in \NRPRRT enhanced the efficiency of both exploration and exploitation in planning.
On average, \NRPRRT required significant less time to achieve per unit of visibility growth of the tree and per unit of distance-to-goal reduction, compared to native RRT. 
Here, visibility is assessed by pre-generating a dense set of global samples and evaluating how many are connectable to the tree, indicating a planner's efficiency in covering unexplored areas. Distance-to-goal, measured as the lowest optimal path length from tree vertices to the goal, reflects a planner's goal-exploitation efficiency.
\NRPRRT demonstrated a notable decrease in required time by approximately 59\% for visibility growth and 54\% for distance-to-goal reduction compared to RRT. 
Results on the required number of vertex expansions further illustrate the learning advantage with \NRPRRT needing 85\% fewer expansions to produce the same unit of visibility growth and distance-to-goal shortening.

\figref{fig:viz_ext_8d} offers a visual comparison of vertex expansion strategies between \NRP and traditional RRT. While RRT's straight-line expansion often resulted in collisions, limiting exploration, \NRP successfully navigated around obstacles using the neural sampler, facilitating successful connection to the target and promoting more efficient tree growth.

\subsection{Conditioning on Local vs. Global Environments}

\begin{table}[t]
\caption{Average optimality score of waypoints produced by locally and globally conditioned samplers in \snake environment. $S_{d}^{local}$ and $S_{g}^{local}$ denote the discriminative sampler and generative sampler in \NRPRRT, and $S_{d}^{global}$ and $S_{g}^{global}$ denote their globally-conditioned counterparts in \NRPRRT-Global.}
\label{table:sampler_performance}
\centering
\begin{footnotesize}
\begin{tabularx}{1\columnwidth}{l@{\extracolsep{\fill}}cc}
    \toprule
    Sampler & Train Environments & Test Environments \\
    \midrule
    $S_{d}^{local}$ & \textcolor{black}{\textbf{0.699}} & \textcolor{black}{\textbf{0.727}} \\
    $S_{g}^{local}$ &  0.599 & 0.619 \\
    $S_{d}^{global}$ & \textcolor{black}{0.230} & \textcolor{black}{0.260} \\
    $S_{g}^{global}$ & 0.217 & 0.143 \\
    \bottomrule
\end{tabularx}
\end{footnotesize}
\vspace{-10pt}
\end{table}

\NRP learns local sampling distribution, allowing the learner to condition on local environments. To analyze the impact of local versus global conditioning on learning efficacy, we contrast \NRPRRT with \NRPRRT-Global, a variant trained with the same dataset but conditions neural samplers on the global environment, $\Wglobal$, instead of $\Wlocal$. The data in \tabref{table:sampler_performance} clearly highlights the advantage of local conditioning, with locally-conditioned samplers enhancing waypoint optimality scores by roughly \textcolor{black}{130\%}. 
The optimality score of a waypoint $q$ is computed as $\frac{L(\solPathOptimal)}{L(\tau(q))}$, where $\solPathOptimal$ denotes the true optimal path and $\tau(q)$ represents the optimal path routing through $q$.
Moreover, locally conditioned samplers excel in test environments, with optimality scores about \textcolor{black}{180\%} higher than globally conditioned ones, achieving strong generalization with no performance drop. In contrast, global samplers' performance degrades in test environments, emphasizing the benefits of local conditioning.

\begin{figure}[t]
    \centering
    \includegraphics[width=\columnwidth]{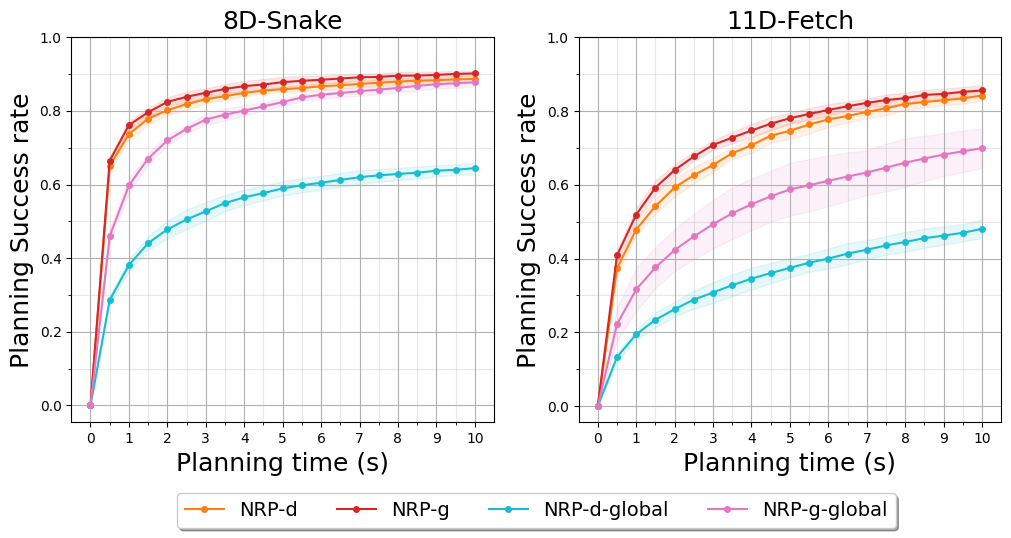}
    \caption{\textcolor{black}{Average planning success rate of NRP when learning with local vs. global conditioning. NRP-d-global and NRP-g-global denote NRP with the globally conditioned discriminative and generative sampler, respectively.}}
    \label{fig:planning_res_local_vs_global}
    \vspace{-10pt}
\end{figure}

\figref{fig:planning_res_local_vs_global} reinforces these findings, showing that local conditioning significantly boosts success rates of planning as \textcolor{black}{\NRPRRT outperforms \NRPRRT-Global in both  \snake and \fetch tasks. Notably, NRP-d achieves a $70\%$ higher success rate than NRP-d-global in \fetch task.}
This enhanced performance demonstrates the effectiveness of learning locally and explains the superior performance of \NRP compared to CVAE and NEXT, which learn globally.

\begin{figure*}[t]
    \centering
    \subfloat[]{\includegraphics[width=0.33\textwidth]{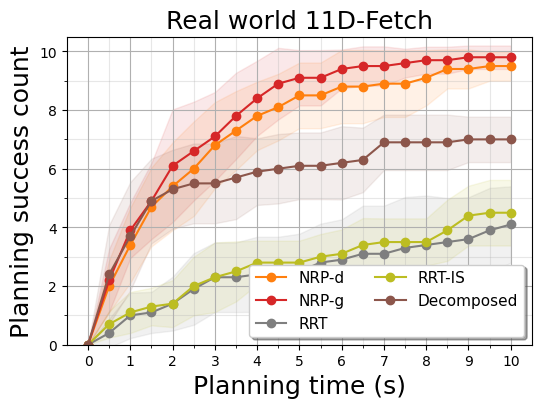}}\hfill
    \subfloat[]{\includegraphics[width=0.33\textwidth]{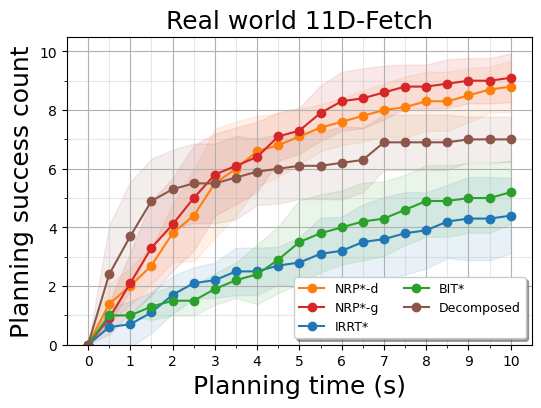}}\hfill
    \subfloat[]{\includegraphics[width=0.33\textwidth]{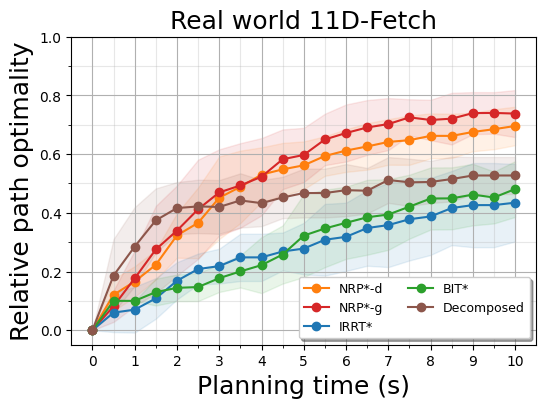}}\hfill
    \caption{\textcolor{black}{Real-world planning performance averaged over 10 runs. (\subfig{a}) Average planning success count of feasible planners. (\subfig{b}) Average planning success count of optimal planners. (\subfig{c}) Average relative path optimality of optimal planners. ``Decomposed'' denotes the domain-specific decomposed planner.}}
    \label{fig:planning_res_rls}
    \vspace{-10pt}
\end{figure*}

\begin{figure}[t]
    \centering
    \subfloat[]{\includegraphics[width=\columnwidth]{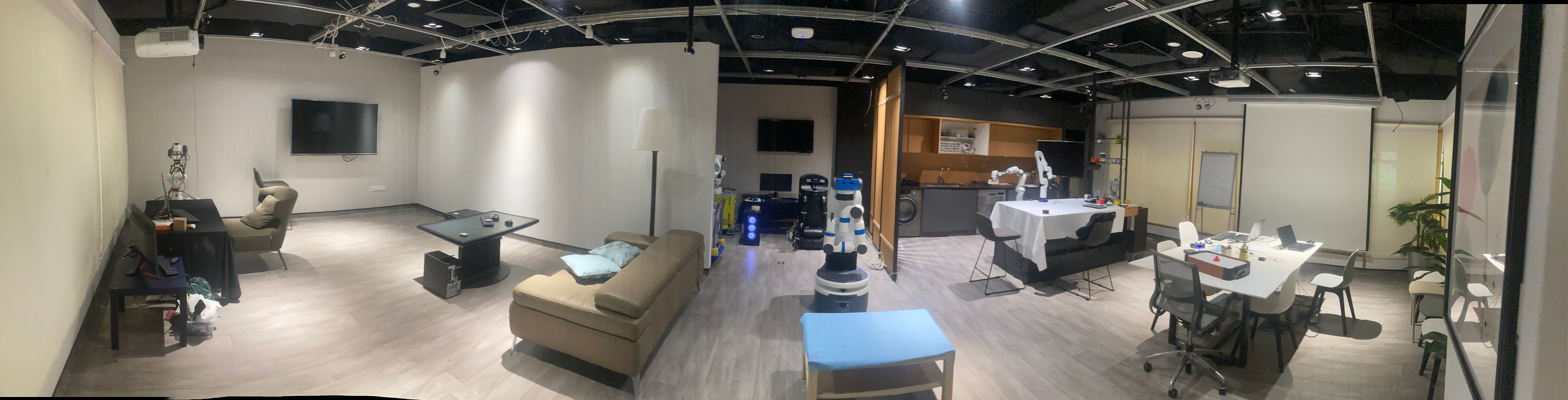}} \hfill
    \subfloat[]{\includegraphics[width=0.33\columnwidth]{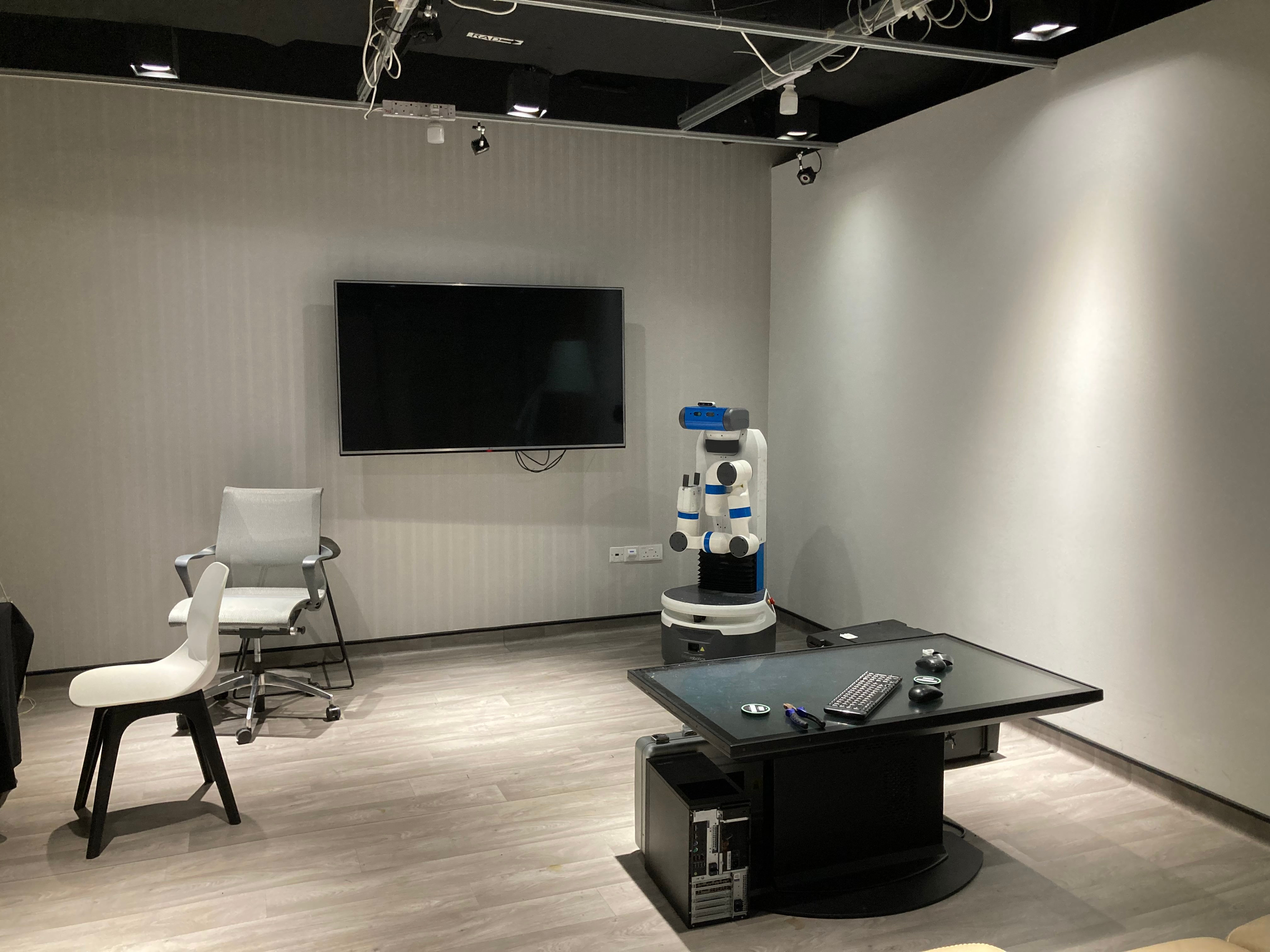}} \hfill
    \subfloat[]{\includegraphics[width=0.33\columnwidth]{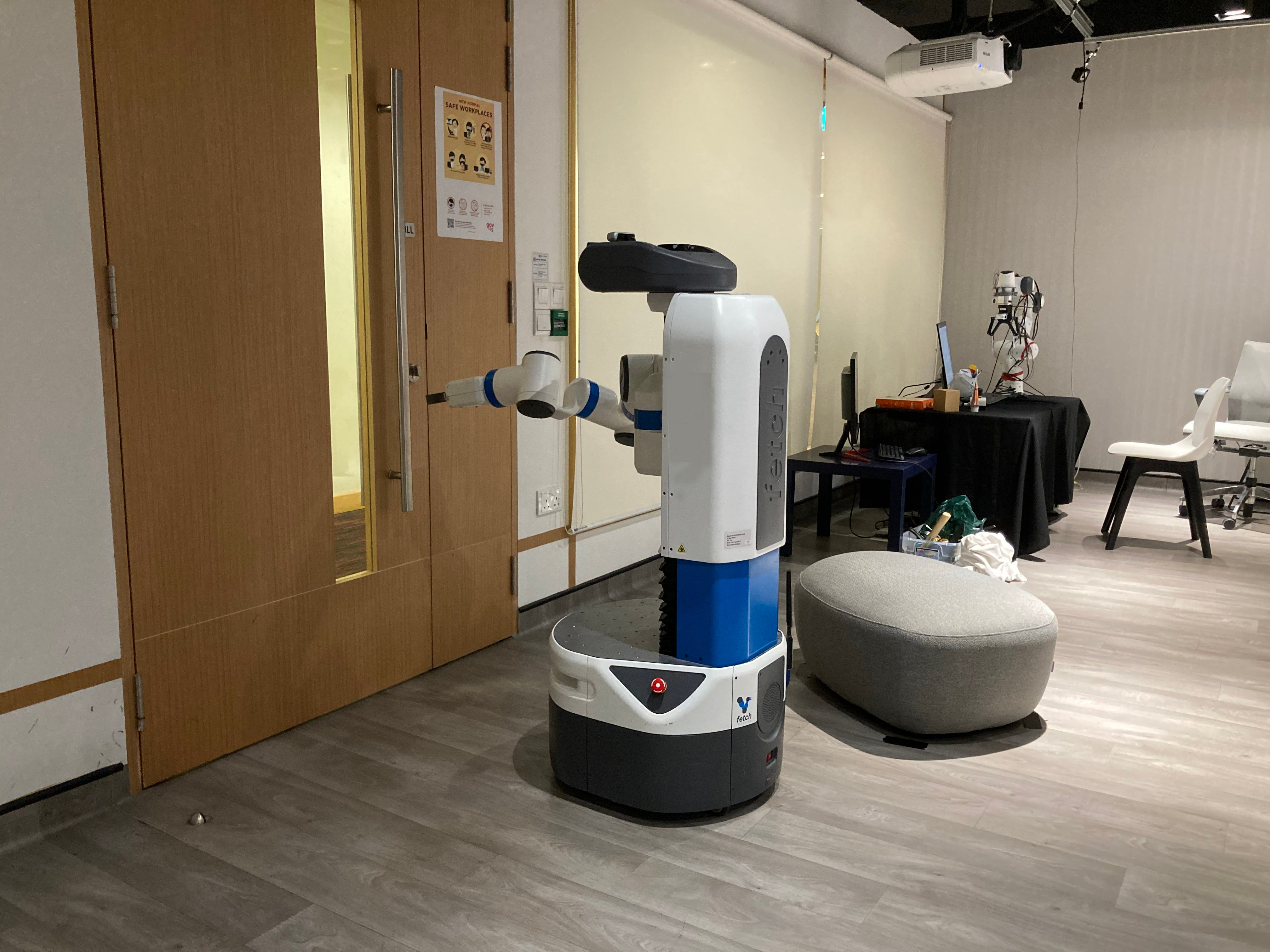}} \hfill
    \subfloat[]{\includegraphics[width=0.33\columnwidth]{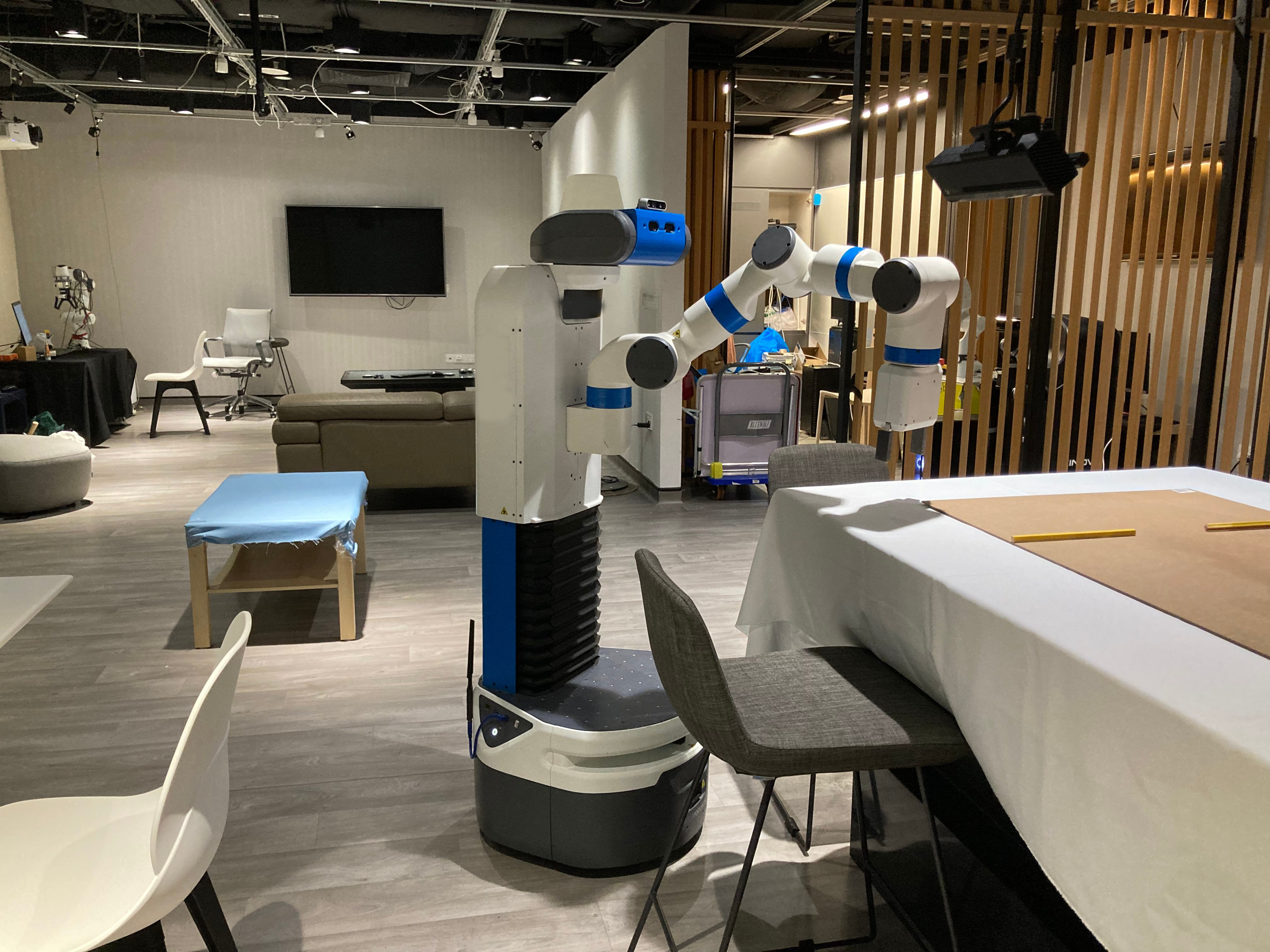}} \hfill
    \newline
    \subfloat[]{\includegraphics[width=0.33\columnwidth]{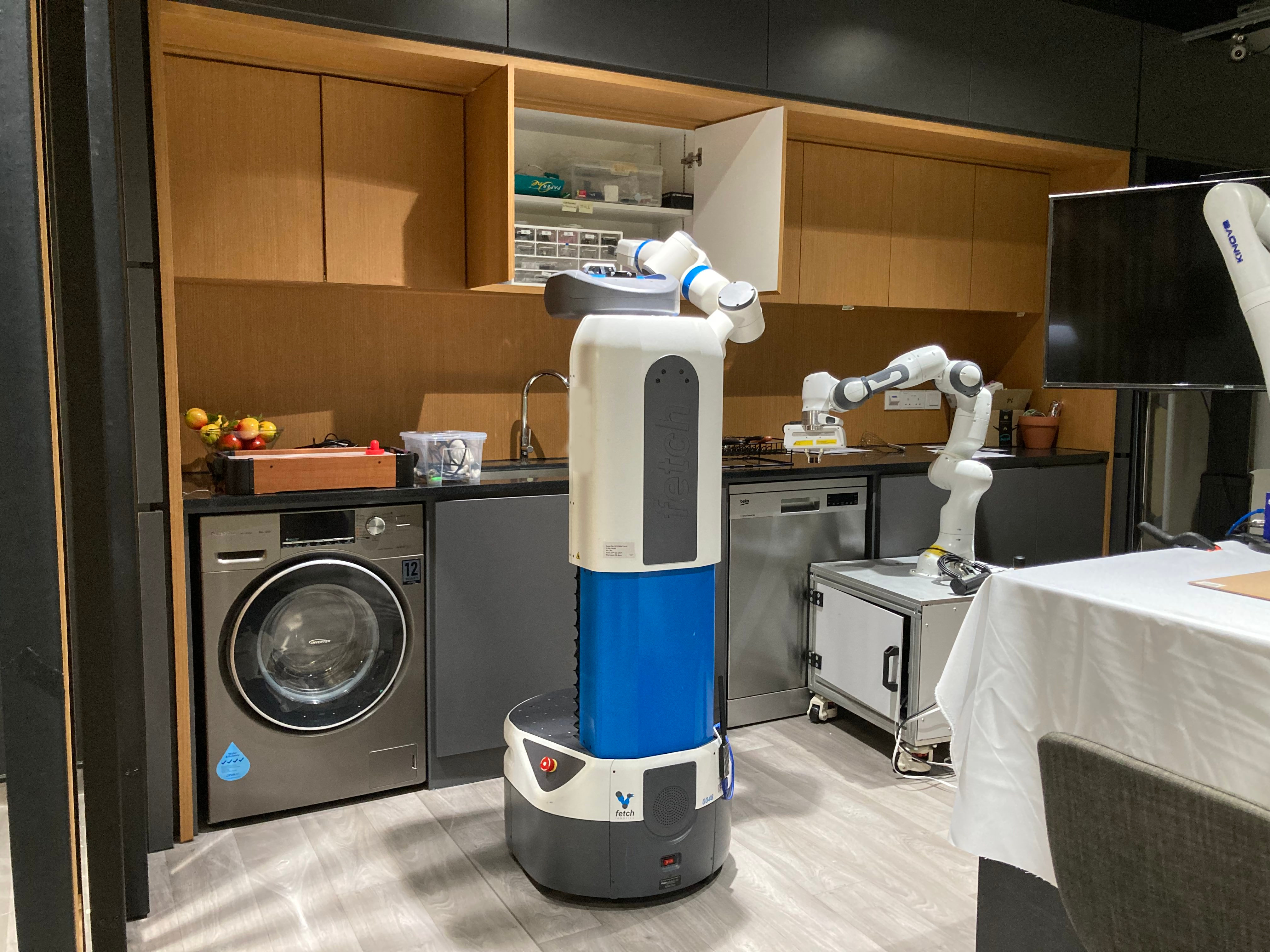}} \hfill
    \subfloat[]{\includegraphics[width=0.33\columnwidth]{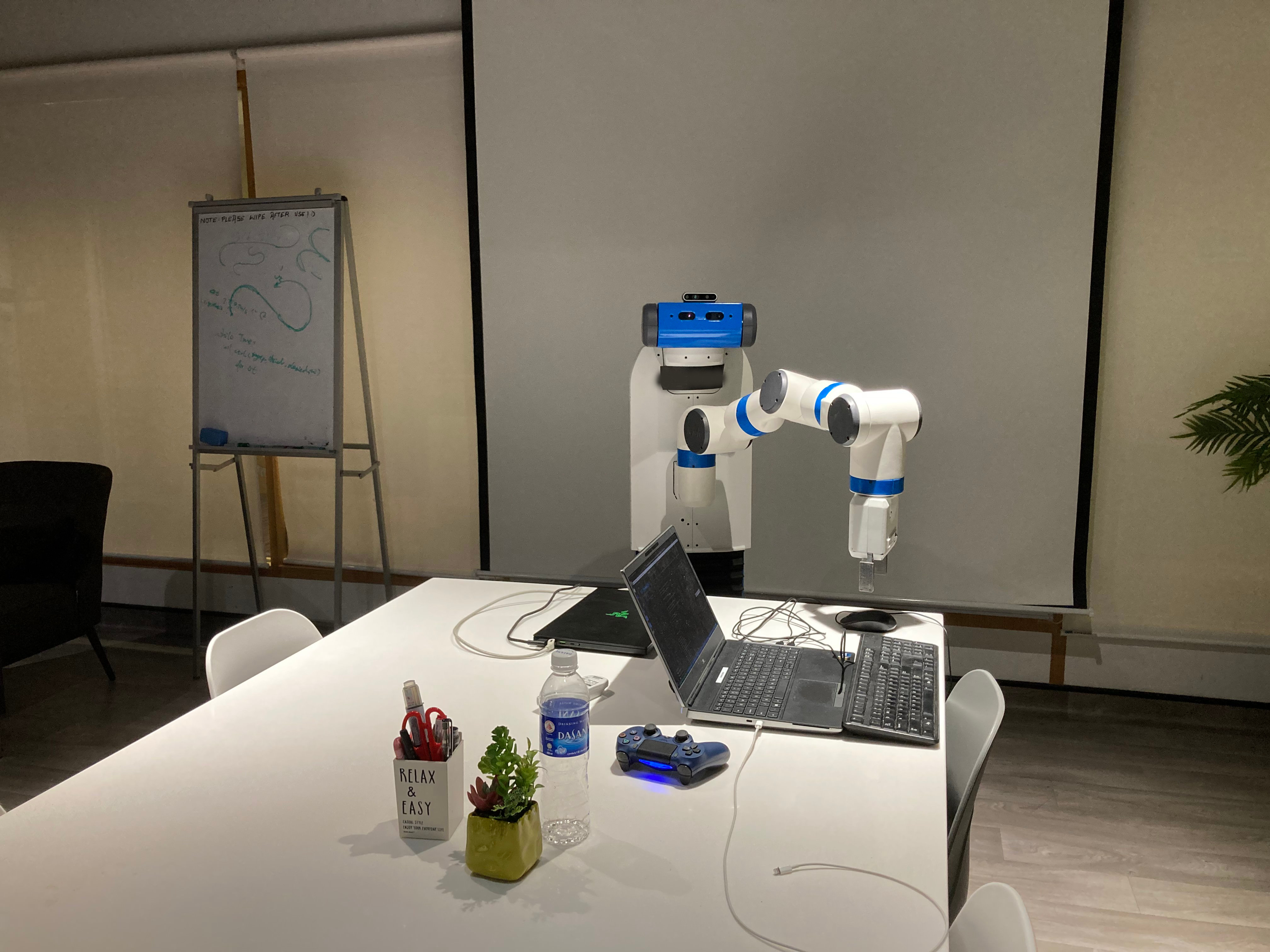}} \hfill
    \subfloat[]{\includegraphics[width=0.33\columnwidth]{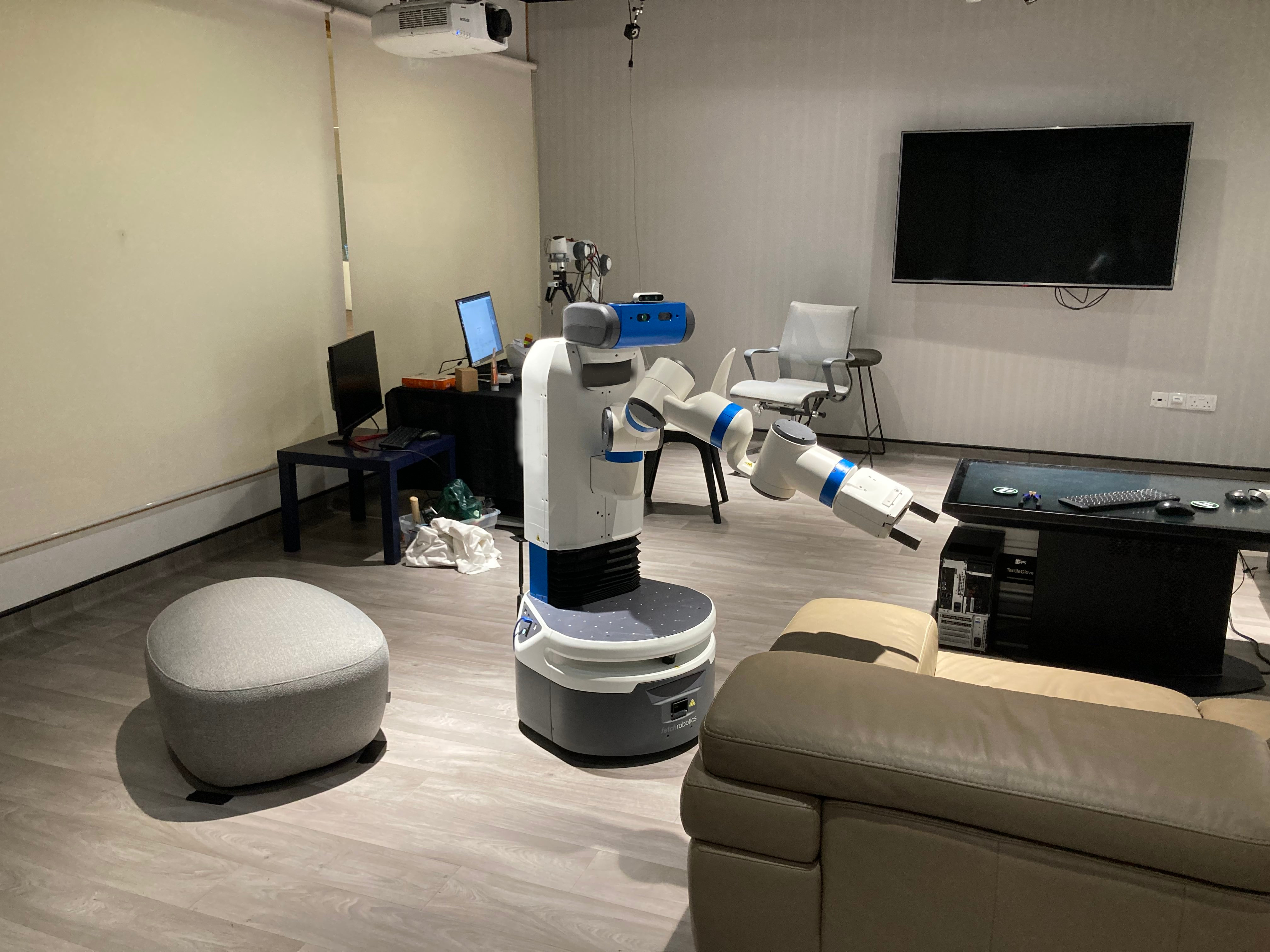}} \hfill
    \caption{The real-world test environment. (\subfig{a}) A panorama view with Fetch robot in the middle. (\subfig{b-g}): Six representative robot configurations: TV, Door, Kitchen, Cabinet, Table and Sofa.}
    \label{fig:rls_configurations}
    \vspace{-10pt}
\end{figure}

\subsection{Discriminative vs. Generative Samplers}

\tabref{table:sampler_performance} also reveals that the discriminative sampler outperformed the generative sampler in producing high-quality samples, evidenced by superior optimality scores in both training and testing environments. 
This performance gap is attributed to the broader learning scope of the discriminative sampler, which benefits from exposure to both optimal and non-optimal samples, ensuring more robust performance. In contrast, the generative sampler only learns from optimal samples and is therefore prone to generating incorrect samples \cite{mousavian20196}.
Interestingly, in the \fetch task, as shown in \figref{fig:planning_res}, \figref{fig:planning_res_optimal}, and \figref{fig:path_cost_optimal}, the narrative shifts. \NRP-g, employing the generative sampler, slightly surpassed \NRP-d, which uses the discriminative sampler, in both success rate and path optimality. This divergence is attributed to the higher execution cost of \NRP-d, which requires first generating of a batch of candidate waypoints. In contrast, the generative sampler directly produces the optimal sample, making it faster in practice.

\section{Real-Robot Experiments} \label{sec:real-exp}
In real-world experiments, we deployed \NRPRRT and \NRPIRRT on a real Fetch robot to reach various configurations in a novel home environment.
Results demonstrated that both \NRPRRT and \NRPIRRT seamlessly transferred into real world without the need of any finetuning and adaptation and outperformed existing SBMP approaches by a significant margin. This highlights the practical utility and robustness of \NRP.


\subsubsection{Experimental setup}

We used a Lidar scanner to scan the environment, producing a 3D occupancy grid representation with a resolution of $0.1m$. 
Six representative configurations in the environment were selected to reflect meaningful poses relevant to daily life tasks (\figref{fig:rls_configurations}). We used these configurations to create 10 test queries, with start and goals configurations randomly sampled from the set.
We directly reuse \NRPRRT and \NRPIRRT trained for simulated \fetch task without any modification.


\subsubsection{Comparison with SBMP algorithms}

Results in \figref{fig:planning_res_rls} show that \NRPRRT and \NRPIRRT significantly outperform existing motion planners ready-to-use for Fetch, including RRT, IRRT*, and BIT*. 
\NRPRRT achieves a \textcolor{black}{$110\%$} higher success rate over the best feasible planer, RRT-IS. \NRPIRRT achieves \textcolor{black}{$80\%$} higher success rate and \textcolor{black}{$40\%$} higher path optimality over the best optimal planner, \textcolor{black}{BIT*}. Notably, \NRP reached $90\%$ success rate after 7 seconds of planning while the baseline algorithms only achieved $40\%$ success rate.

\subsubsection{Comparison with domain-specific planner}

\textcolor{black}{Additionally, we compared our approach with a decomposed planning approach that involves an initial phase using IRRT* to position the robot arm into a resting configuration, followed by employing Hybrid-A* \cite{dolgov2008practical} to navigate the robot base towards the goal location, and concluding with another application of IRRT* to extend the arm to the goal configuration. This decomposed planner tackles the whole-body planning challenge by manually decomposing the problem into lower-dimensional ones. This hand-crafted strategy initially permits slightly higher success rates and path optimality over \NRPIRRT, as depicted in \figref{fig:planning_res_rls}. However, the lack of whole-body coordination leads to a loss in asymptotic completeness and optimality, resulting in quickly plateaued performance over time. In contrast, \NRPIRRT demonstrates superior asymptotic performance, achieving about 28\% higher success rate and 35\% higher path optimality.}

\textcolor{black}{\tabref{table:real_comparison} further compares the execution time of paths produced by \NRPIRRT and the decomposed planner for a long-range task involving 5 sub-queries. Both planners used 10 seconds of planning time in this study.
Our observations indicate that \NRPIRRT significantly enhances operational efficiency, roughly doubling the speed of task completion compared to the decomposed planner. The paths generated by \NRPIRRT, as documented in the accompanying video, demonstrate smoother and more naturally coordinated whole-body motion, optimizing overall task accomplishment.
}

\begin{table}[t]
\caption{Execution time of paths generated by \NRPIRRT and decomposed planning. Both methods use 10 seconds to plan each query.}
\label{table:real_comparison}
\centering
\begin{footnotesize}
\begin{tabularx}{1\columnwidth}{l@{\extracolsep{\fill}}cc}
    \toprule
    \textbf{Query} & \multicolumn{2}{c}{Path Execution Time (s)}\\
    \cmidrule(lr){2-3}
    & \NRPIRRT & Decomposed planning \\
    \midrule
    TV to Door & 23 & 44 \\
    Door to Kitchen & 33 & 44 \\
    Kitchen to Cabinet & 21 & 43 \\
    Cabinet to Table & 33 & 68 \\
    Table to Sofa & 40 & 80 \\
    Total & 141 & 279 \\
    \bottomrule
\end{tabularx}
\end{footnotesize}
\vspace{-10pt}
\end{table}

\section{Conclusion}
This paper introduced the Neural Randomized Planner (\NRP), a novel algorithm designed to enhance whole-body motion planning for high-DOF robots in household environments through learning local sampling distributions. Our experimental results demonstrated that \NRP outperforms both classical and contemporary learning-based SBMP algorithms in complex planning tasks, and showcased its effectiveness in both simulations and real-world applications. Despite its success, \NRP assumes a fully known and static global environment, a condition not always met in some practical scenarios where robots have to deal with partially-known spaces and dynamic obstacles. An intriguing avenue for future research involves adapting \NRP to achieve fast re-planning, enabling the system to dynamically adjust to new information and obstacles, thereby extending its applicability to more realistic and challenging environments. \textcolor{black}{Furthermore, NRP only considers primitive  neural networks that might not be expressive enough to capture extreme multi-model distributions. More advanced architectures, such as diffusion, could be explored to enhance the quality of learned sampling distribution.}

\bibliographystyle{plainnat}
\bibliography{references}

\begin{thebibliography}{63}
\providecommand{\natexlab}[1]{#1}
\providecommand{\url}[1]{\texttt{#1}}
\expandafter\ifx\csname urlstyle\endcsname\relax
  \providecommand{\doi}[1]{doi: #1}\else
  \providecommand{\doi}{doi: \begingroup \urlstyle{rm}\Url}\fi

\bibitem[Atreya and Biswas(2022)]{atreya2022state}
Pranav Atreya and Joydeep Biswas.
\newblock State supervised steering function for sampling-based kinodynamic planning.
\newblock In \emph{Proceedings of the 21st International Conference on Autonomous Agents and Multiagent Systems}, AAMAS '22, page 35–43, Richland, SC, 2022. International Foundation for Autonomous Agents and Multiagent Systems.
\newblock ISBN 9781450392136.

\bibitem[Burget et~al.(2013)Burget, Hornung, and Bennewitz]{burget2013whole}
Felix Burget, Armin Hornung, and Maren Bennewitz.
\newblock Whole-body motion planning for manipulation of articulated objects.
\newblock In \emph{2013 IEEE International Conference on Robotics and Automation}, pages 1656--1662. IEEE, 2013.

\bibitem[Castaman et~al.(2021)Castaman, Pagello, Menegatti, and Pretto]{castaman2021receding}
Nicola Castaman, Enrico Pagello, Emanuele Menegatti, and Alberto Pretto.
\newblock Receding horizon task and motion planning in changing environments.
\newblock \emph{Robotics and Autonomous Systems}, 145:\penalty0 103863, 2021.

\bibitem[Chamzas et~al.(2019)Chamzas, Shrivastava, and Kavraki]{chamzas2019using}
Constantinos Chamzas, Anshumali Shrivastava, and Lydia~E Kavraki.
\newblock Using local experiences for global motion planning.
\newblock In \emph{2019 International Conference on Robotics and Automation (ICRA)}, pages 8606--8612. IEEE, 2019.

\bibitem[Chamzas et~al.(2021)Chamzas, Kingston, Quintero-Pe{\~n}a, Shrivastava, and Kavraki]{chamzas2021learning}
Constantinos Chamzas, Zachary Kingston, Carlos Quintero-Pe{\~n}a, Anshumali Shrivastava, and Lydia~E Kavraki.
\newblock Learning sampling distributions using local 3d workspace decompositions for motion planning in high dimensions.
\newblock In \emph{2021 IEEE International Conference on Robotics and Automation (ICRA)}, pages 1283--1289. IEEE, 2021.

\bibitem[Chamzas et~al.(2022)Chamzas, Cullen, Shrivastava, and Kavraki]{chamzas2022learning}
Constantinos Chamzas, Aedan Cullen, Anshumali Shrivastava, and Lydia~E Kavraki.
\newblock Learning to retrieve relevant experiences for motion planning.
\newblock In \emph{2022 International Conference on Robotics and Automation (ICRA)}, pages 7233--7240. IEEE, 2022.

\bibitem[Chen et~al.(2019)Chen, Dai, Lin, Ye, Liu, and Song]{chen2019learning}
Binghong Chen, Bo~Dai, Qinjie Lin, Guo Ye, Han Liu, and Le~Song.
\newblock Learning to plan in high dimensions via neural exploration-exploitation trees.
\newblock \emph{arXiv preprint arXiv:1903.00070}, 2019.

\bibitem[Cheng et~al.(2020)Cheng, Shankar, and Burdick]{cheng2020learning}
Richard Cheng, Krishna Shankar, and Joel~W. Burdick.
\newblock Learning an optimal sampling distribution for efficient motion planning.
\newblock In \emph{2020 IEEE/RSJ International Conference on Intelligent Robots and Systems (IROS)}, pages 7485--7492, 2020.
\newblock \doi{10.1109/IROS45743.2020.9341245}.

\bibitem[Chitta et~al.(2010)Chitta, Cohen, and Likhachev]{chitta2010planning}
Sachin Chitta, Benjamin Cohen, and Maxim Likhachev.
\newblock Planning for autonomous door opening with a mobile manipulator.
\newblock In \emph{2010 IEEE International Conference on Robotics and Automation}, pages 1799--1806. IEEE, 2010.

\bibitem[Chitta et~al.(2012)Chitta, Jones, Ciocarlie, and Hsiao]{chitta2012mobile}
Sachin Chitta, E~Gil Jones, Matei Ciocarlie, and Kaijen Hsiao.
\newblock Mobile manipulation in unstructured environments: Perception, planning, and execution.
\newblock \emph{IEEE Robotics \& Automation Magazine}, 19\penalty0 (2):\penalty0 58--71, 2012.

\bibitem[Dai et~al.(2014)Dai, Valenzuela, and Tedrake]{dai2014whole}
Hongkai Dai, Andr{\'e}s Valenzuela, and Russ Tedrake.
\newblock Whole-body motion planning with centroidal dynamics and full kinematics.
\newblock In \emph{2014 IEEE-RAS International Conference on Humanoid Robots}, pages 295--302. IEEE, 2014.

\bibitem[Das and Yip(2020)]{das2020learning}
Nikhil Das and Michael Yip.
\newblock Learning-based proxy collision detection for robot motion planning applications.
\newblock \emph{IEEE Transactions on Robotics}, 36\penalty0 (4):\penalty0 1096--1114, 2020.

\bibitem[Dolgov et~al.(2008)Dolgov, Thrun, Montemerlo, and Diebel]{dolgov2008practical}
Dmitri Dolgov, Sebastian Thrun, Michael Montemerlo, and James Diebel.
\newblock Practical search techniques in path planning for autonomous driving.
\newblock \emph{Ann Arbor}, 1001\penalty0 (48105):\penalty0 18--80, 2008.

\bibitem[Elbanhawi and Simic(2014)]{elbanhawi2014sampling}
Mohamed Elbanhawi and Milan Simic.
\newblock Sampling-based robot motion planning: A review.
\newblock \emph{Ieee access}, 2:\penalty0 56--77, 2014.

\bibitem[Fu et~al.(2023)Fu, Cheng, and Pathak]{fu2023deep}
Zipeng Fu, Xuxin Cheng, and Deepak Pathak.
\newblock Deep whole-body control: learning a unified policy for manipulation and locomotion.
\newblock In \emph{Conference on Robot Learning}, pages 138--149. PMLR, 2023.

\bibitem[Gammell et~al.(2014)Gammell, Srinivasa, and Barfoot]{gammell2014informed}
Jonathan~D Gammell, Siddhartha~S Srinivasa, and Timothy~D Barfoot.
\newblock Informed rrt*: Optimal sampling-based path planning focused via direct sampling of an admissible ellipsoidal heuristic.
\newblock In \emph{2014 IEEE/RSJ International Conference on Intelligent Robots and Systems}, pages 2997--3004. IEEE, 2014.

\bibitem[Gammell et~al.(2015)Gammell, Srinivasa, and Barfoot]{gammell2015batch}
Jonathan~D Gammell, Siddhartha~S Srinivasa, and Timothy~D Barfoot.
\newblock Batch informed trees (bit): Sampling-based optimal planning via the heuristically guided search of implicit random geometric graphs.
\newblock In \emph{2015 IEEE international conference on robotics and automation (ICRA)}, pages 3067--3074. IEEE, 2015.

\bibitem[Haddad et~al.(2006)Haddad, Chettibi, Hanchi, and Lehtihet]{haddad2006optimal}
M~Haddad, T~Chettibi, S~Hanchi, and HE~Lehtihet.
\newblock Optimal motion planner of mobile manipulators in generalized point-to-point task.
\newblock In \emph{9th IEEE International Workshop on Advanced Motion Control, 2006.}, pages 300--306. IEEE, 2006.

\bibitem[Honerkamp et~al.(2023)Honerkamp, Welschehold, and Valada]{honerkamp2023n}
Daniel Honerkamp, Tim Welschehold, and Abhinav Valada.
\newblock $\textrm{N}^2 \textrm{M}^2 $: Learning navigation for arbitrary mobile manipulation motions in unseen and dynamic environments.
\newblock \emph{IEEE Transactions on Robotics}, 2023.

\bibitem[Hsu et~al.(2006)Hsu, Latombe, and Kurniawati]{hsu2006probabilistic}
David Hsu, Jean-Claude Latombe, and Hanna Kurniawati.
\newblock On the probabilistic foundations of probabilistic roadmap planning.
\newblock \emph{The International Journal of Robotics Research}, 25\penalty0 (7):\penalty0 627--643, 2006.

\bibitem[Hu et~al.(2023)Hu, Stone, and Mart{\'\i}n-Mart{\'\i}n]{hu2023causal}
Jiaheng Hu, Peter Stone, and Roberto Mart{\'\i}n-Mart{\'\i}n.
\newblock Causal policy gradient for whole-body mobile manipulation.
\newblock \emph{arXiv preprint arXiv:2305.04866}, 2023.

\bibitem[Huh and Lee(2018)]{huh2018efficient}
Jinwook Huh and Daniel~D. Lee.
\newblock Efficient sampling with q-learning to guide rapidly exploring random trees.
\newblock \emph{IEEE Robotics and Automation Letters}, 3\penalty0 (4):\penalty0 3868--3875, 2018.
\newblock \doi{10.1109/LRA.2018.2856927}.

\bibitem[Ichter et~al.(2018)Ichter, Harrison, and Pavone]{ichter2018learning}
Brian Ichter, James Harrison, and Marco Pavone.
\newblock Learning sampling distributions for robot motion planning.
\newblock In \emph{2018 IEEE International Conference on Robotics and Automation (ICRA)}, pages 7087--7094. IEEE, 2018.

\bibitem[Ichter et~al.(2020)Ichter, Schmerling, Lee, and Faust]{ichter2020learned}
Brian Ichter, Edward Schmerling, Tsang-Wei~Edward Lee, and Aleksandra Faust.
\newblock Learned critical probabilistic roadmaps for robotic motion planning.
\newblock In \emph{2020 IEEE International Conference on Robotics and Automation (ICRA)}, pages 9535--9541. IEEE, 2020.

\bibitem[Jauhri et~al.(2022)Jauhri, Peters, and Chalvatzaki]{jauhri2022robot}
Snehal Jauhri, Jan Peters, and Georgia Chalvatzaki.
\newblock Robot learning of mobile manipulation with reachability behavior priors.
\newblock \emph{IEEE Robotics and Automation Letters}, 7\penalty0 (3):\penalty0 8399--8406, 2022.

\bibitem[Jurgenson and Tamar(2019)]{jurgenson2019harnessing}
Tom Jurgenson and Aviv Tamar.
\newblock Harnessing reinforcement learning for neural motion planning.
\newblock \emph{arXiv preprint arXiv:1906.00214}, 2019.

\bibitem[Karaman and Frazzoli(2011)]{karaman2011sampling}
Sertac Karaman and Emilio Frazzoli.
\newblock Sampling-based algorithms for optimal motion planning.
\newblock \emph{The international journal of robotics research}, 30\penalty0 (7):\penalty0 846--894, 2011.

\bibitem[Kew et~al.(2019)Kew, Ichter, Bandari, Lee, and Faust]{kew2019neural}
J~Chase Kew, Brian Ichter, Maryam Bandari, Tsang-Wei~Edward Lee, and Aleksandra Faust.
\newblock Neural collision clearance estimator for batched motion planning.
\newblock \emph{arXiv preprint arXiv:1910.05917}, 2019.

\bibitem[Khan et~al.(2020)Khan, Ribeiro, Kumar, and Francis]{khan2020graph}
Arbaaz Khan, Alejandro Ribeiro, Vijay Kumar, and Anthony~G Francis.
\newblock Graph neural networks for motion planning.
\newblock \emph{arXiv preprint arXiv:2006.06248}, 2020.

\bibitem[Kicki and Skrzypczyński(2022)]{kicki2022speeding}
Piotr Kicki and Piotr Skrzypczyński.
\newblock Speeding up deep neural network-based planning of local car maneuvers via efficient b-spline path construction.
\newblock In \emph{2022 International Conference on Robotics and Automation (ICRA)}, pages 4422--4428, 2022.
\newblock \doi{10.1109/ICRA46639.2022.9812313}.

\bibitem[Kicki et~al.(2021)Kicki, Gawron, Ćwian, Ozay, and Skrzypczyński]{kicki2021learning}
Piotr Kicki, Tomasz Gawron, Krzysztof Ćwian, Mete Ozay, and Piotr Skrzypczyński.
\newblock Learning from experience for rapid generation of local car maneuvers.
\newblock \emph{Engineering Applications of Artificial Intelligence}, 105:\penalty0 104399, 2021.
\newblock ISSN 0952-1976.
\newblock \doi{https://doi.org/10.1016/j.engappai.2021.104399}.
\newblock URL \url{https://www.sciencedirect.com/science/article/pii/S0952197621002475}.

\bibitem[Kicki et~al.(2024)Kicki, Liu, Tateo, Bou-Ammar, Walas, Skrzypczyński, and Peters]{kicki2024fast}
Piotr Kicki, Puze Liu, Davide Tateo, Haitham Bou-Ammar, Krzysztof Walas, Piotr Skrzypczyński, and Jan Peters.
\newblock Fast kinodynamic planning on the constraint manifold with deep neural networks.
\newblock \emph{IEEE Transactions on Robotics}, 40:\penalty0 277--297, 2024.
\newblock \doi{10.1109/TRO.2023.3326922}.

\bibitem[Kobashi et~al.(2023)Kobashi, Wang, Zhao, Lin, and Tomizuka]{kobashi2023learning}
Keita Kobashi, Changhao Wang, Yu~Zhao, Hsien-Chung Lin, and Masayoshi Tomizuka.
\newblock Learning from local experience: Informed sampling distributions for high dimensional motion planning, 2023.

\bibitem[Kumar et~al.(2019)Kumar, Mandalika, Choudhury, and Srinivasa]{kumar2019lego}
Rahul Kumar, Aditya Mandalika, Sanjiban Choudhury, and Siddhartha Srinivasa.
\newblock Lego: Leveraging experience in roadmap generation for sampling-based planning.
\newblock In \emph{2019 IEEE/RSJ International Conference on Intelligent Robots and Systems (IROS)}, pages 1488--1495. IEEE, 2019.

\bibitem[LaValle et~al.(1998)]{lavalle1998rapidly}
Steven~M LaValle et~al.
\newblock Rapidly-exploring random trees: A new tool for path planning.
\newblock 1998.

\bibitem[Lembono et~al.(2021)Lembono, Pignat, Jankowski, and Calinon]{lembono2021learning}
Teguh~Santoso Lembono, Emmanuel Pignat, Julius Jankowski, and Sylvain Calinon.
\newblock Learning constrained distributions of robot configurations with generative adversarial network, 2021.

\bibitem[Li et~al.(2020)Li, Xia, Martin-Martin, and Savarese]{li2020hrl4in}
Chengshu Li, Fei Xia, Roberto Martin-Martin, and Silvio Savarese.
\newblock Hrl4in: Hierarchical reinforcement learning for interactive navigation with mobile manipulators.
\newblock In \emph{Conference on Robot Learning}, pages 603--616. PMLR, 2020.

\bibitem[Li et~al.(2021)Li, Miao, Qureshi, and Yip]{li2021mpc}
Linjun Li, Yinglong Miao, Ahmed~H. Qureshi, and Michael~C. Yip.
\newblock Mpc-mpnet: Model-predictive motion planning networks for fast, near-optimal planning under kinodynamic constraints.
\newblock \emph{IEEE Robotics and Automation Letters}, 6\penalty0 (3):\penalty0 4496--4503, 2021.
\newblock \doi{10.1109/LRA.2021.3067847}.

\bibitem[Luna et~al.(2020)Luna, Moll, Badger, and Kavraki]{luna2020scalable}
Ryan Luna, Mark Moll, Julia Badger, and Lydia~E Kavraki.
\newblock A scalable motion planner for high-dimensional kinematic systems.
\newblock \emph{The International Journal of Robotics Research}, 39\penalty0 (4):\penalty0 361--388, 2020.

\bibitem[McMahon et~al.(2022)McMahon, Sivaramakrishnan, Granados, Bekris, et~al.]{mcmahon2022survey}
Troy McMahon, Aravind Sivaramakrishnan, Edgar Granados, Kostas~E Bekris, et~al.
\newblock A survey on the integration of machine learning with sampling-based motion planning.
\newblock \emph{Foundations and Trends{\textregistered} in Robotics}, 9\penalty0 (4):\penalty0 266--327, 2022.

\bibitem[Molina et~al.(2020)Molina, Kumar, and Srivastava]{molina2020learn}
Daniel Molina, Kislay Kumar, and Siddharth Srivastava.
\newblock Learn and link: Learning critical regions for efficient planning.
\newblock In \emph{2020 IEEE International Conference on Robotics and Automation (ICRA)}, pages 10605--10611. IEEE, 2020.

\bibitem[Mousavian et~al.(2019)Mousavian, Eppner, and Fox]{mousavian20196}
Arsalan Mousavian, Clemens Eppner, and Dieter Fox.
\newblock 6-dof graspnet: Variational grasp generation for object manipulation.
\newblock In \emph{Proceedings of the IEEE/CVF International Conference on Computer Vision}, pages 2901--2910, 2019.

\bibitem[Qureshi et~al.(2019)Qureshi, Simeonov, Bency, and Yip]{qureshi2019motion}
Ahmed~H Qureshi, Anthony Simeonov, Mayur~J Bency, and Michael~C Yip.
\newblock Motion planning networks.
\newblock In \emph{2019 International Conference on Robotics and Automation (ICRA)}, pages 2118--2124. IEEE, 2019.

\bibitem[Qureshi et~al.(2020)Qureshi, Dong, Choe, and Yip]{qureshi2020neural}
Ahmed~H Qureshi, Jiangeng Dong, Austin Choe, and Michael~C Yip.
\newblock Neural manipulation planning on constraint manifolds.
\newblock \emph{IEEE Robotics and Automation Letters}, 5\penalty0 (4):\penalty0 6089--6096, 2020.

\bibitem[Qureshi et~al.(2022)Qureshi, Dong, Baig, and Yip]{qureshi2022constrained}
Ahmed~Hussain Qureshi, Jiangeng Dong, Asfiya Baig, and Michael~C. Yip.
\newblock Constrained motion planning networks x.
\newblock \emph{IEEE Transactions on Robotics}, 38\penalty0 (2):\penalty0 868--886, 2022.
\newblock \doi{10.1109/TRO.2021.3096070}.

\bibitem[Rastegarpanah et~al.(2021)Rastegarpanah, Gonzalez, and Stolkin]{rastegarpanah2021semi}
Alireza Rastegarpanah, Hector~Cruz Gonzalez, and Rustam Stolkin.
\newblock Semi-autonomous behaviour tree-based framework for sorting electric vehicle batteries components.
\newblock \emph{Robotics}, 10\penalty0 (2):\penalty0 82, 2021.

\bibitem[Reeds and Shepp(1990)]{reeds1990optimal}
James Reeds and Lawrence Shepp.
\newblock Optimal paths for a car that goes both forwards and backwards.
\newblock \emph{Pacific journal of mathematics}, 145\penalty0 (2):\penalty0 367--393, 1990.

\bibitem[Saoji and Rosell(2020)]{saoji2020flexibly}
Siddhant Saoji and Jan Rosell.
\newblock Flexibly configuring task and motion planning problems for mobile manipulators.
\newblock In \emph{2020 25th IEEE International Conference on Emerging Technologies and Factory Automation (ETFA)}, volume~1, pages 1285--1288. IEEE, 2020.

\bibitem[Schramm and Boularias(2022)]{schramm2022learning}
Liam Schramm and Abdeslam Boularias.
\newblock Learning-guided exploration for efficient sampling-based motion planning in high dimensions.
\newblock In \emph{2022 International Conference on Robotics and Automation (ICRA)}, pages 4429--4435. IEEE, 2022.

\bibitem[Sohn et~al.(2015)Sohn, Lee, and Yan]{sohn2015learning}
Kihyuk Sohn, Honglak Lee, and Xinchen Yan.
\newblock Learning structured output representation using deep conditional generative models.
\newblock \emph{Advances in neural information processing systems}, 28, 2015.

\bibitem[Stilman(2010)]{stilman2010global}
Mike Stilman.
\newblock Global manipulation planning in robot joint space with task constraints.
\newblock \emph{IEEE Transactions on Robotics}, 26\penalty0 (3):\penalty0 576--584, 2010.

\bibitem[Stilman et~al.(2007)Stilman, Schamburek, Kuffner, and Asfour]{stilman2007manipulation}
Mike Stilman, Jan-Ullrich Schamburek, James Kuffner, and Tamim Asfour.
\newblock Manipulation planning among movable obstacles.
\newblock In \emph{Proceedings 2007 IEEE international conference on robotics and automation}, pages 3327--3332. IEEE, 2007.

\bibitem[Strudel et~al.(2021)Strudel, Pinel, Carpentier, Laumond, Laptev, and Schmid]{strudel2021learning}
Robin Strudel, Ricardo~Garcia Pinel, Justin Carpentier, Jean-Paul Laumond, Ivan Laptev, and Cordelia Schmid.
\newblock Learning obstacle representations for neural motion planning.
\newblock In \emph{Conference on Robot Learning}, pages 355--364. PMLR, 2021.

\bibitem[Tran et~al.(2020)Tran, Denny, and Ekenna]{tran2020predicting}
Tuan Tran, Jory Denny, and Chinwe Ekenna.
\newblock Predicting sample collision with neural networks.
\newblock \emph{arXiv preprint arXiv:2006.16868}, 2020.

\bibitem[Urmson and Simmons(2003)]{urmson2003approaches}
Chris Urmson and Reid Simmons.
\newblock Approaches for heuristically biasing rrt growth.
\newblock In \emph{Proceedings 2003 IEEE/RSJ International Conference on Intelligent Robots and Systems (IROS 2003)(Cat. No. 03CH37453)}, volume~2, pages 1178--1183. IEEE, 2003.

\bibitem[Wang et~al.(2020)Wang, Chi, Li, Wang, and Meng]{wang2020neural}
Jiankun Wang, Wenzheng Chi, Chenming Li, Chaoqun Wang, and Max Q-H Meng.
\newblock Neural rrt*: Learning-based optimal path planning.
\newblock \emph{IEEE Transactions on Automation Science and Engineering}, 17\penalty0 (4):\penalty0 1748--1758, 2020.

\bibitem[Wei et~al.(2019)Wei, Jiang, Rahmani, and Zhan]{wei2019motion}
Yan Wei, Wei Jiang, Ahmed Rahmani, and Qiang Zhan.
\newblock Motion planning for a humanoid mobile manipulator system.
\newblock \emph{International Journal of Humanoid Robotics}, 16\penalty0 (02):\penalty0 1950006, 2019.

\bibitem[Wolfslag et~al.(2018)Wolfslag, Bharatheesha, Moerland, and Wisse]{wolfslag2018rrt}
Wouter~J. Wolfslag, Mukunda Bharatheesha, Thomas~M. Moerland, and Martijn Wisse.
\newblock Rrt-colearn: Towards kinodynamic planning without numerical trajectory optimization.
\newblock \emph{IEEE Robotics and Automation Letters}, 3\penalty0 (3):\penalty0 1655--1662, 2018.
\newblock \doi{10.1109/LRA.2018.2801470}.

\bibitem[Xia et~al.(2018)Xia, Zamir, He, Sax, Malik, and Savarese]{xia2018gibson}
Fei Xia, Amir~R Zamir, Zhiyang He, Alexander Sax, Jitendra Malik, and Silvio Savarese.
\newblock Gibson env: Real-world perception for embodied agents.
\newblock In \emph{Proceedings of the IEEE conference on computer vision and pattern recognition}, pages 9068--9079, 2018.

\bibitem[Xia et~al.(2021)Xia, Li, Mart{\'\i}n-Mart{\'\i}n, Litany, Toshev, and Savarese]{xia2021relmogen}
Fei Xia, Chengshu Li, Roberto Mart{\'\i}n-Mart{\'\i}n, Or~Litany, Alexander Toshev, and Silvio Savarese.
\newblock Relmogen: Integrating motion generation in reinforcement learning for mobile manipulation.
\newblock In \emph{2021 IEEE International Conference on Robotics and Automation (ICRA)}, pages 4583--4590. IEEE, 2021.

\bibitem[Yavari et~al.(2019)Yavari, Gupta, and Mehrandezh]{yavari2019lazy}
Mohammadreza Yavari, Kamal Gupta, and Mehran Mehrandezh.
\newblock Lazy steering rrt*: An optimal constrained kinodynamic neural network based planner with no in-exploration steering.
\newblock In \emph{2019 19th International Conference on Advanced Robotics (ICAR)}, pages 400--407, 2019.
\newblock \doi{10.1109/ICAR46387.2019.8981551}.

\bibitem[Yu and Gao(2021)]{yu2021reducing}
Chenning Yu and Sicun Gao.
\newblock Reducing collision checking for sampling-based motion planning using graph neural networks.
\newblock \emph{Advances in Neural Information Processing Systems}, 34:\penalty0 4274--4289, 2021.

\bibitem[Zhang et~al.(2018)Zhang, Huh, and Lee]{zhang2018learning}
Clark Zhang, Jinwook Huh, and Daniel~D Lee.
\newblock Learning implicit sampling distributions for motion planning.
\newblock In \emph{2018 IEEE/RSJ International Conference on Intelligent Robots and Systems (IROS)}, pages 3654--3661. IEEE, 2018.

\end{thebibliography}

\section{Appendix}

\begin{figure}[t]
    \centering
    \includegraphics[width=\columnwidth]{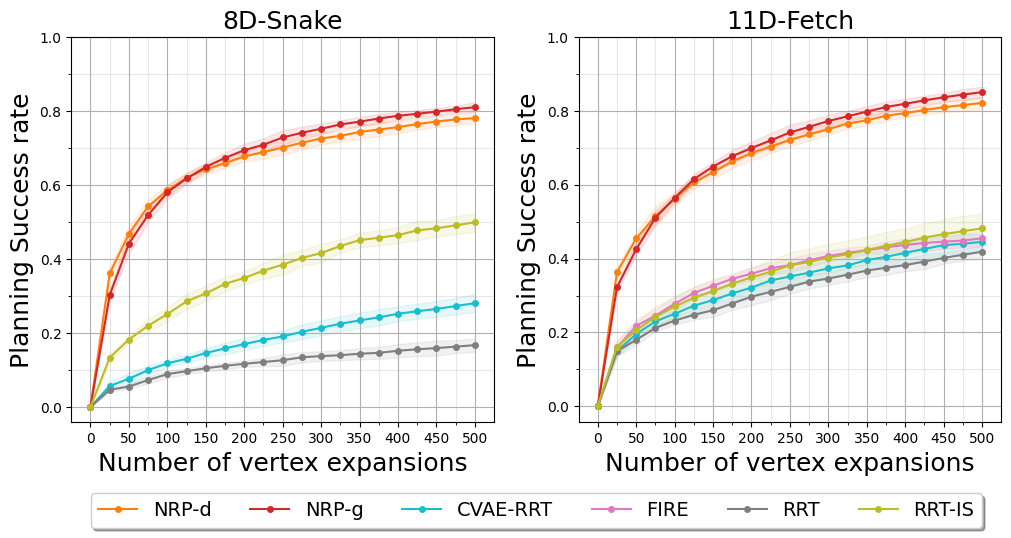}
    \caption{\textcolor{black}{Average planning success rate for feasible planners.}}
    \label{fig:planning_res}
    \vspace{-10pt}
\end{figure}

\begin{figure}[t]
    \centering
    \includegraphics[width=\columnwidth]{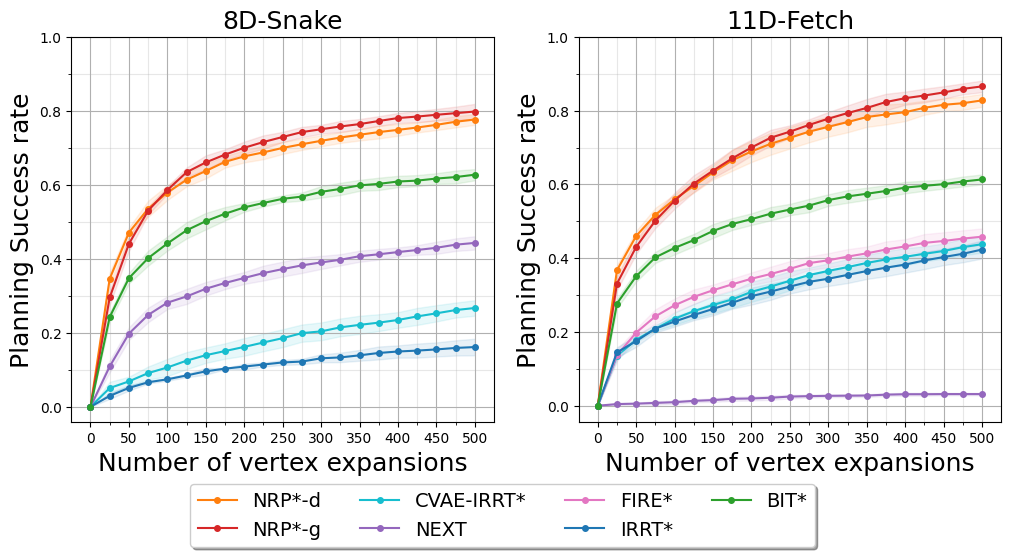}
    \caption{\textcolor{black}{Average planning success rate for optimal planners.}}
    \label{fig:planning_res_optimal}
    \vspace{-10pt}
\end{figure}

\begin{figure}[t]
    \centering
    \includegraphics[width=\columnwidth]{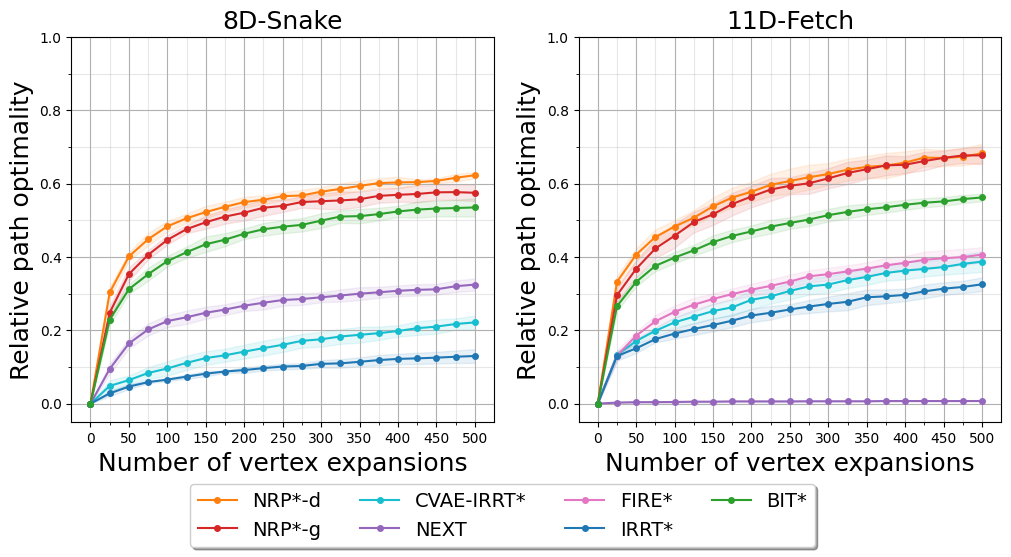}
    \caption{\textcolor{black}{Average relative path optimality for optimal planners.}}
    \label{fig:path_cost_optimal}
\end{figure}

\begin{figure*}[t]
    \subfloat[]{\includegraphics[width=0.49\textwidth]{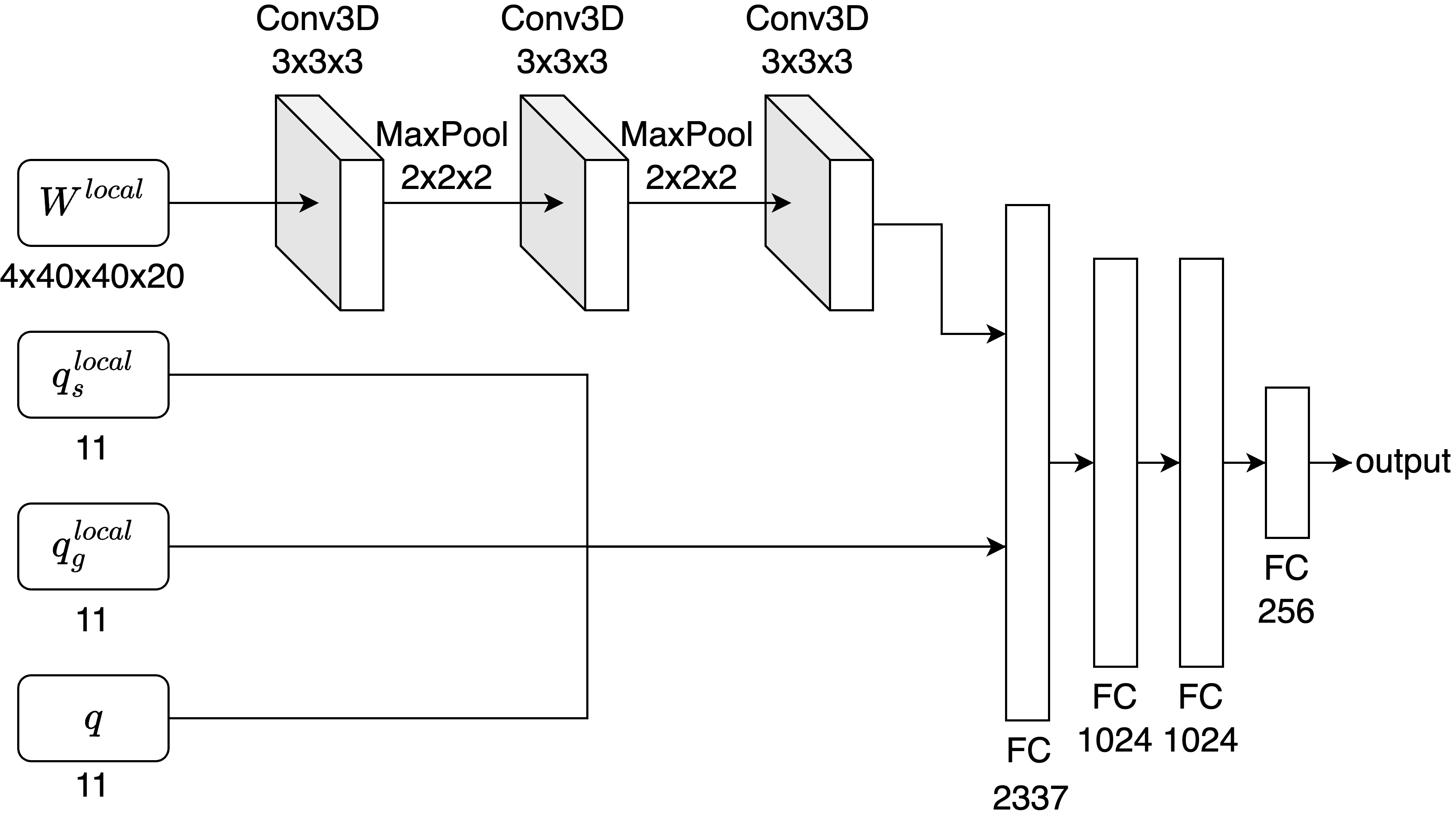}}\hfill
    \subfloat[]{\includegraphics[width=0.49\textwidth]{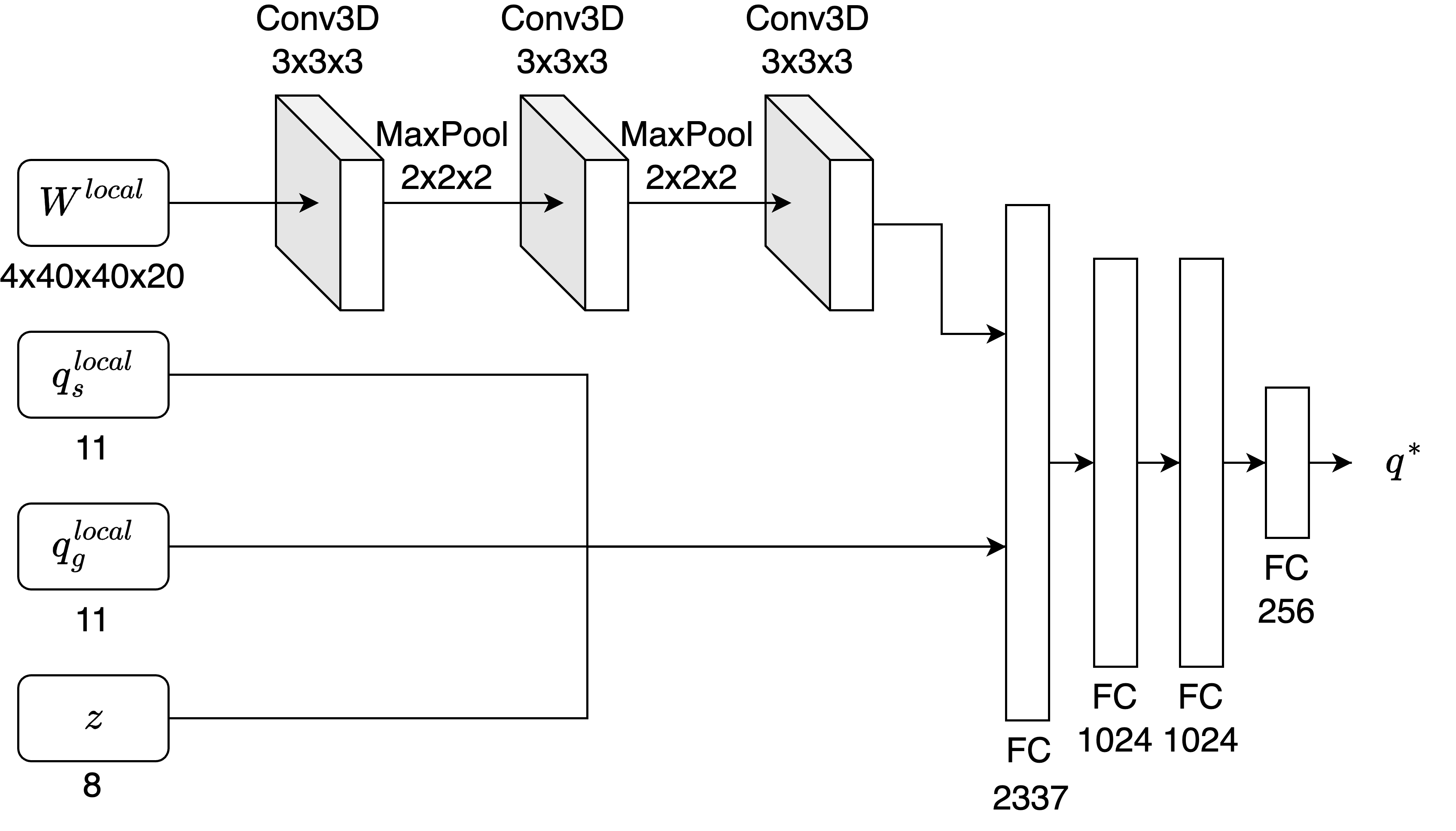}}\hfill
    \caption{\NRP's network architecture under \fetch task. (\subfig{a}) \textcolor{black}{Network architecture of $\optNetwork$ and CVAE encoder. The two networks have different outputs. Specifically, $\optNetwork$ outputs probability $p_{q}^{opt}$, and encoder outputs mean and variance of the latent distribution.} (\subfig{b}) Network architecture of CVAE decoder where z is a sample in the latent space and $q^{*}$ is the reconstructed optimal waypoint. In both figures, Conv3D denotes a 3D convolutional layer. MaxPool denotes the max pooling layer which takes the maximum value out of every subgrid. FC denotes a fully connected layer. The activation function for all the layers is ReLU. The size of each layer are shown in the figure. $\Wlocal$ is represented by 3D occupancy grids with 4 channels, formed by each voxel's occupied status and its x,y,z cartesian coordinates. }
    \label{fig:network_architecture}

\end{figure*}

\begin{figure*}[t]
    \centering
    \hfill
    \subfloat[] {\includegraphics[width=0.33\textwidth]{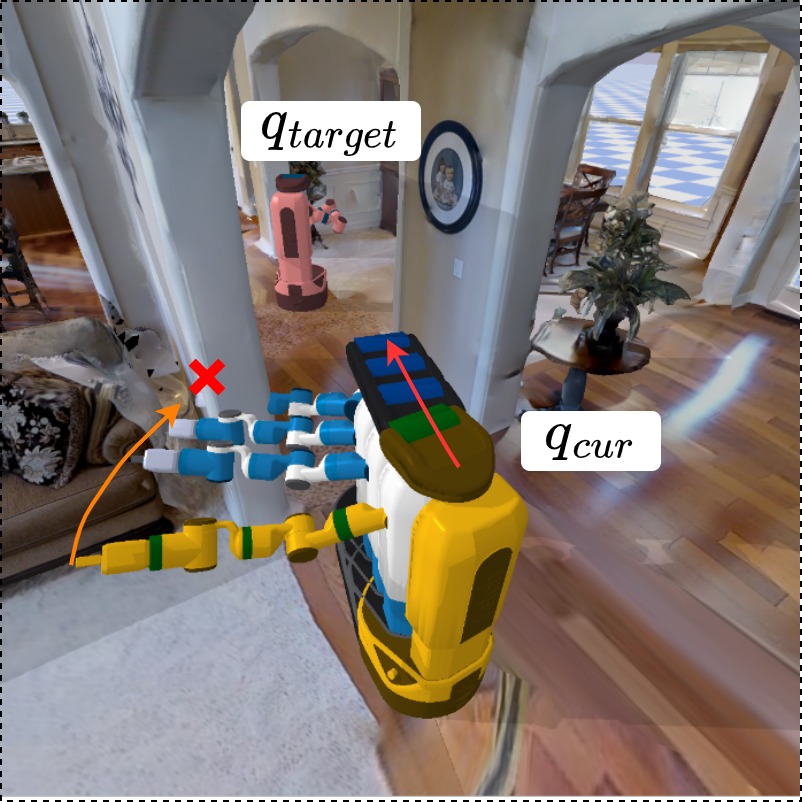}}\hfill
    \subfloat[]{\includegraphics[width=0.33\textwidth]{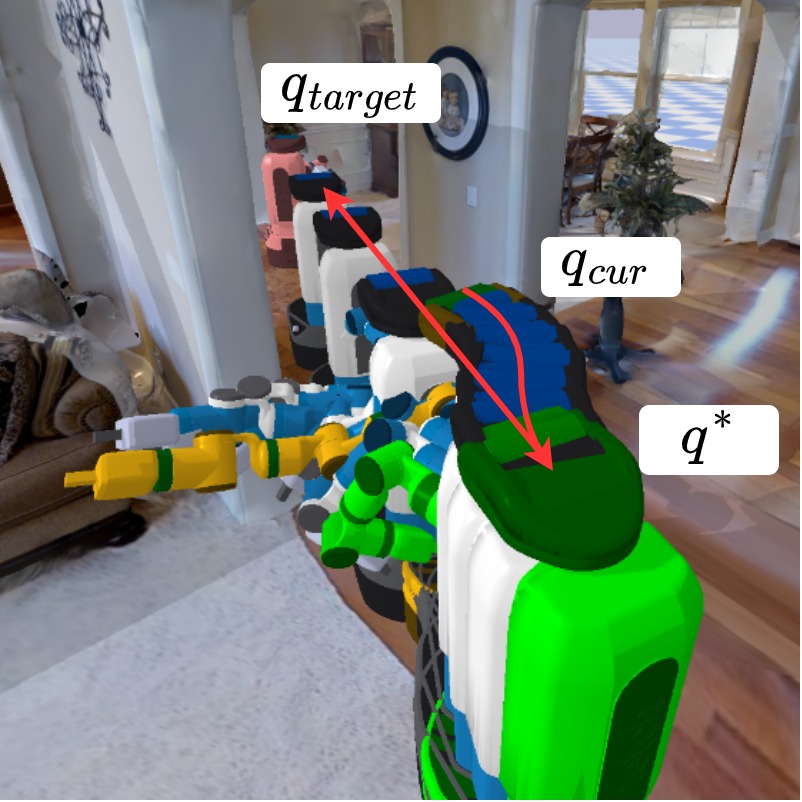}}\hfill
    \subfloat[]{\includegraphics[width=0.33\textwidth]{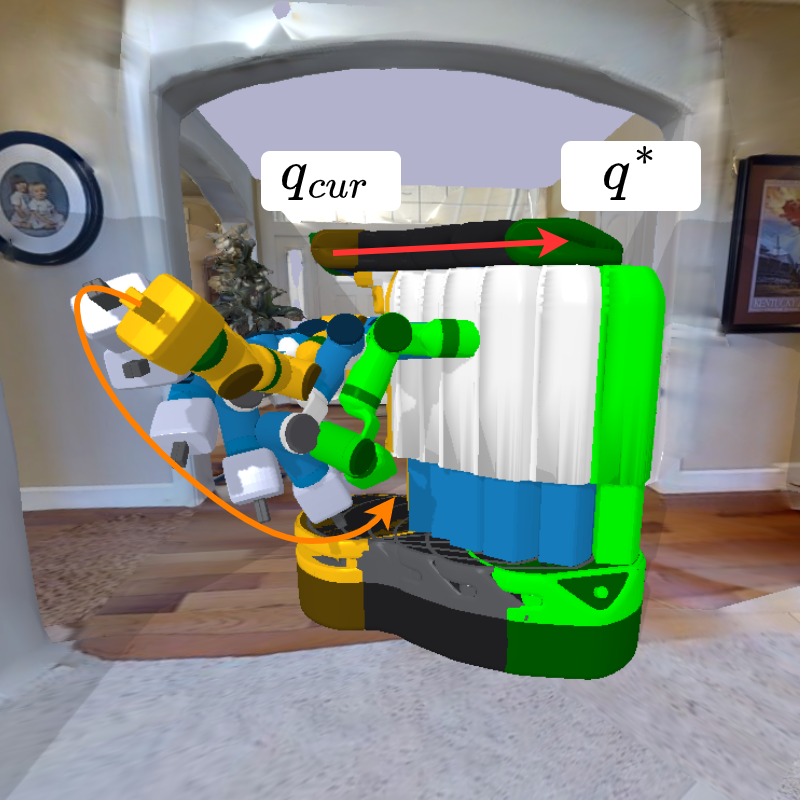}}\hfill
    \caption{\textcolor{black}{Visualization of vertex expansion path in the \fetch task. The Fetch robot's base trajectory is drawn with a red arrow and its end-effector trajectory is drawn with brown arrow. 
    (\subfig{a}) Straight-line expansion in RRT attempted to move robot straight from $\qcur$ (yellow) towards $\qtarget$ (red), causing arm to quickly collide with walls. (\subfig{b}) Neural sampler in \NRP produced a waypoint $\optimalWaypoint$ (green) that moved the robot backward first, giving enough room for robot to tuck its arm into a conservative configuration, allowing the robot to subsequently move towards $\qtarget$ in a straight-line along the narrow corridor. (\subfig{c}) Side view of the Fetch robot tucking its arm following the guidance of $\optimalWaypoint$.}}
    \label{fig:viz_ext_11d}
    \vspace{-10pt}
\end{figure*}

\subsection{Planning performance with respect to number of expansions}
In this section, the planning performance of \NRP was compared against baselines with equal numbers of vertex expansions allowed, in both \snake and \fetch task. Unlike planning time which is influenced by the hardware setup (\eg CPU and GPU speed), assessing performance based on the number of vertex expansions provides more insights into the algorithmic efficiency of planners.

Results in \figref{fig:planning_res}, \figref{fig:planning_res_optimal} and \figref{fig:path_cost_optimal} demonstrate that \NRP consistently produced significantly higher planning success rate and path optimality than all baselines. 
In the context of feasible planning, \NRP achieved a success rate approximately 30\% higher than RRT-IS, the best classical planner, and 50\% higher than CVAE-RRT, the best learning-enhanced planner, in both \snake and \fetch tasks. 
In the context of optimal planning, \NRP attained a success rate around 20\% higher than BIT*, the best classical planner, and 30-40\% higher than NEXT and CVAE-IRRT*, the best learning-enhanced planners in \snake and \fetch tasks, respectively. 
This superior performance underscores the efficacy of learned sampling distributions in \NRP.

\subsection{Hyperparameters of \NRP}\label{sec:appendix}

\begin{table}[t]
\caption{Optimal hyperparameters for \NRP to achieve the best planning performance against allowed planning time. }
\label{table:hyperparameter_8d}
\centering
\begin{tabularx}{1\columnwidth}{l@{\extracolsep{\fill}}cc}
    \toprule
    Algorithm & Goal bias & Shortest-path expansion rate \\
    \midrule
    RRT/IRRT* & 0.1 & -  \\
    \NRPRRT-d/\NRPIRRT-d  & 0.5 & 0.2 \\
    \NRPRRT-g/\NRPIRRT-g  & 0.4 & 0.2 \\
    \bottomrule
\end{tabularx} 
\end{table}

\tabref{table:hyperparameter_8d} presents the hyperparameter used for \NRP, compared to RRT and IRRT*. 
An interesting finding is that \NRP achieved optimal performance with a larger goal bias compared to original RRT and IRRT*. Goal bias, controlling the probability of sampling the global goal as the vertex expansion target, balances the exploitation and exploration of the planner. A higher goal bias indicates \NRP can afford to be more greedy, as its neural sampler leads to effective exploration. In contrast, original RRT and IRRT* need a lower goal bias, investing more effort in exploration to compensate for their less efficient shortest-path expansion.

Additionally, we determined the optimal shortest-path expansion rate, the rate at which \NRP should perform shortest-path expansion instead of using the neural sampler. Result indicates that a relatively low rate of 0.2 achieved the best planning performance, proving that \NRP's neural sampler is effective and should be favoured in most of the cases. 

\subsection{Network architecture of NRP}\label{sec:network_architecture}

All networks used by \NRP, including $\optNetwork$ in the discriminative sampler, and encoder, decoder in the generative sampler, share similar network architecture, illustrated by \figref{fig:network_architecture}. 

Specifically, the occupancy grid representation of local environment $\Wlocal$ is first passed through a series of convolution layers to obtain an 1D feature vector. For \fetch task with 3D occupancy grids, we use 3D convolution layers. For \snake task with 2D occupancy grids, we use 2D convolution layers. This feature vector is concatenated together with other inputs to create the full latent vector, which is then passed through a series of fully-connected (FC) layers to produce the final output. For \fetch task, FC layers have a size of 1024. For \snake task, FC layers have a size of 512.




\subsection{\textcolor{black}{Additional Qualitative Analysis of Neural Expansion}}\label{sec:qualitative_neural_expansion}
\textcolor{black}{This section provides further visualization of expansion paths generated by \NRP in the \fetch task, highlighting a particularly challenging scenario. Here, the robot must intricately coordinate its base and arm to reach $\qtarget$, as depicted in \figref{fig:viz_ext_11d}. Although the base positions at $\qcur$ and $\qtarget$ could be connected by a simple straight-line, such a path would lead to collisions between the arm and surrounding obstacles. To address this, \NRP suggests $\optimalWaypoint$, a strategic waypoint that moves Fetch backward while maintaining a direct line passage for the base to $\qtarget$, and in the meantime, maneuvering the robot's arm into a safer, collision-free configuration. The resulting expansion path not only avoids obstacles but also significantly narrows the gap to $\qtarget$. This example highlights \NRP’s ability to effectively generate whole-body motions that address complex local planning challenges.}
 
\subsection{\textcolor{black}{Visualization of training and test environments}}
\begin{figure*}[h]
    \centering
    \subfloat[] {\includegraphics[width=0.49\textwidth]{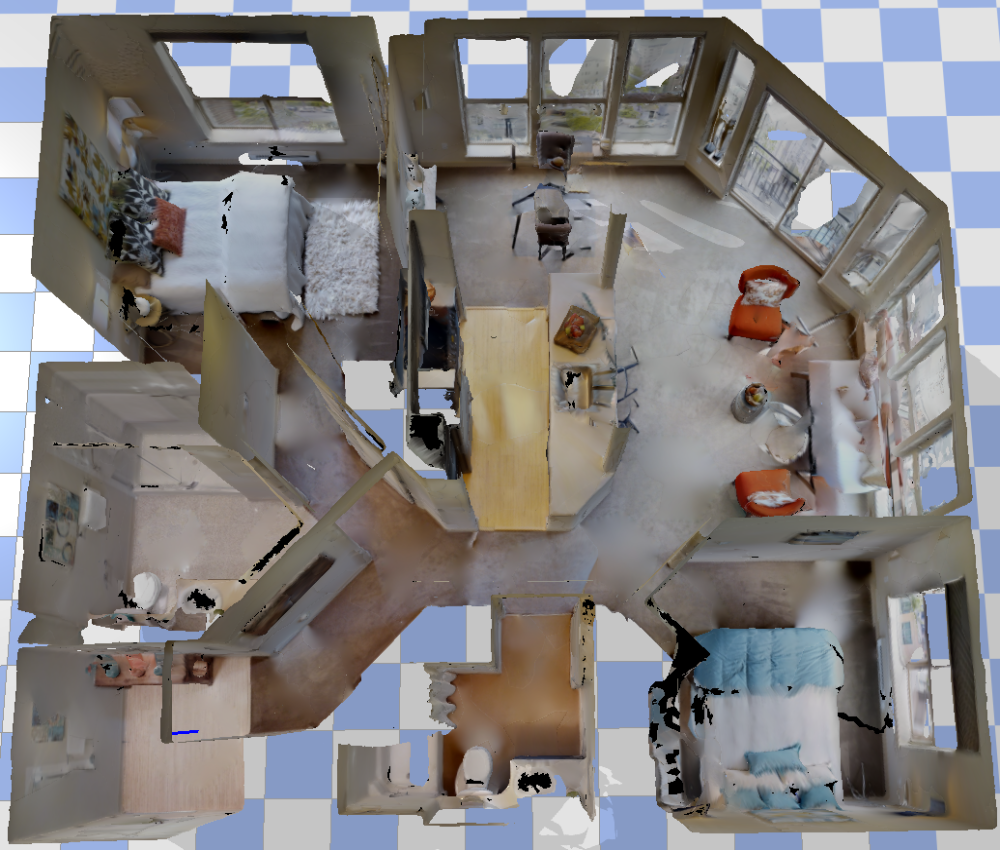}}\hfill
    \subfloat[]{\includegraphics[width=0.49\textwidth]{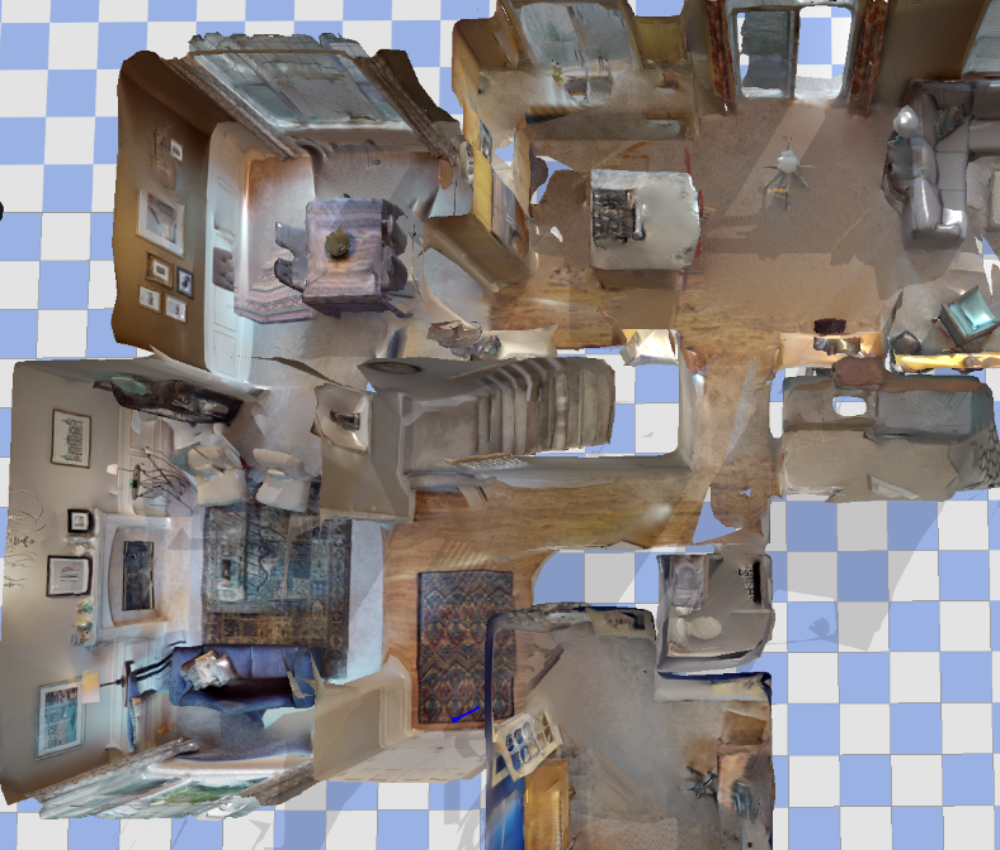}}\hfill
    \subfloat[]{\includegraphics[width=0.49\textwidth]{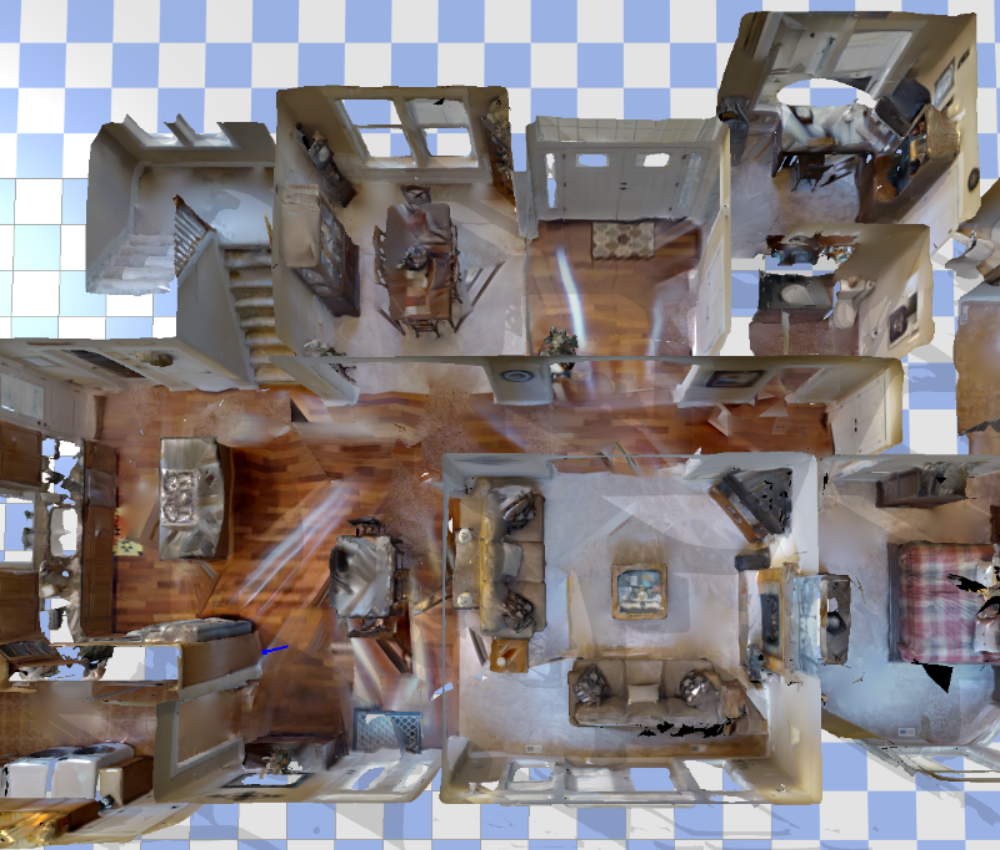}}\hfill
    \subfloat[]{\includegraphics[width=0.49\textwidth]{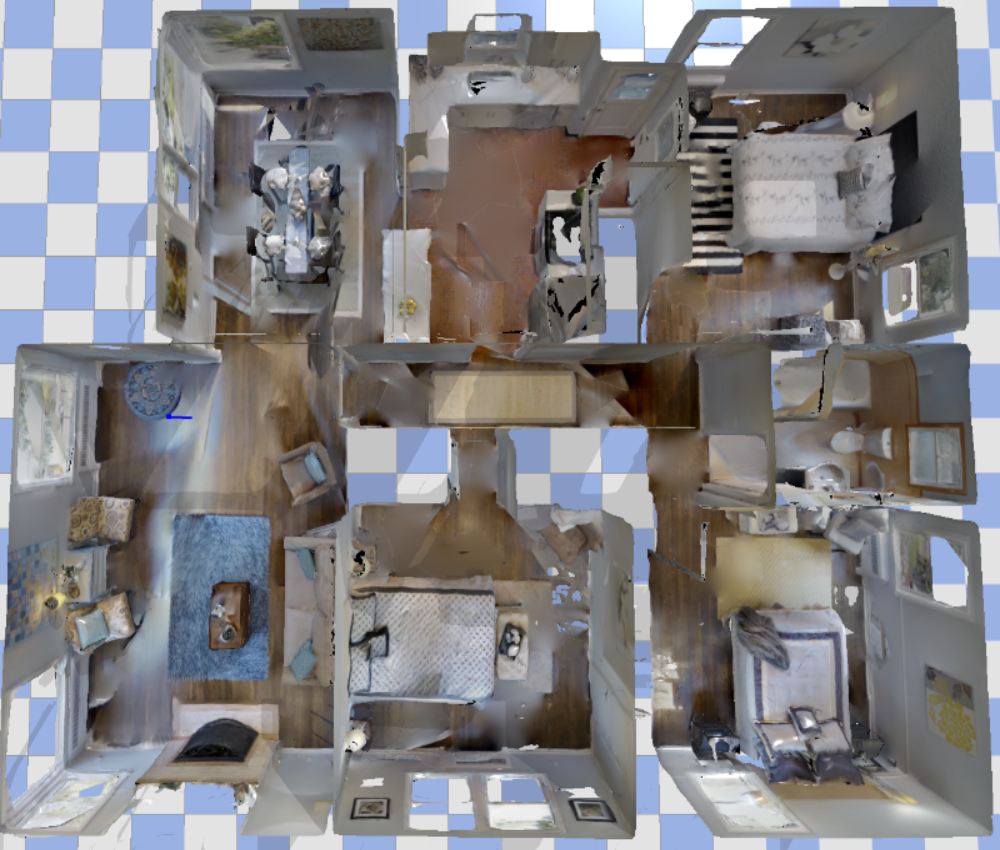}}\hfill
    \subfloat[]{\includegraphics[width=\textwidth]{figure/rls/rls.jpg}}\hfill
    \caption{\textcolor{black}{Examples of training and test environments used in \fetch task. In addition to the simulation environments, \NRP was evaluated in an unseen real household environment in the zero-shot manner, without any fine tuning and adaptation.
    (\subfig{a, b}) Simulation training environments. (\subfig{c, d}) Simulation test environments. (\subfig{e}) Real-world test environment.}}
    \label{fig:env_viz_11d}
    \vspace{-10pt}
\end{figure*}

\textcolor{black}{
\figref{fig:env_viz_11d} shows a set of exemplary training and test environments used in \fetch task. They differ drastically in global layout, geometric details, and visual appearance, which significantly increases the generalization difficulty for methods that directly model a global sampling distribution or a global policy.
}

\end{document}